%% file: main.tex
\documentclass[manuscript, preprint, nonacm]{acmart}
\AtBeginDocument{%
  }

\usepackage[T1]{fontenc}
\usepackage[utf8]{inputenc}

\usepackage{listings}

\usepackage{ulem}
\usepackage{hyperref}
\usepackage{enumitem}
\usepackage{geometry}
\usepackage{graphicx} 
\usepackage{subcaption}
\usepackage{array}
\usepackage[most]{tcolorbox}
\usepackage[table]{xcolor}
\usepackage{booktabs}
\usepackage{tabularx}
\usepackage{multirow}

\tcbset{
    academicexample/.style={
        colback=gray!10,
        colframe=gray!70,
        sharp corners,
        boxrule=0.8pt,
        fontupper=\small,
        left=8pt, right=8pt,
        top=6pt, bottom=6pt,
        width=\linewidth
    }
}

\begin{document}

\title{A Taxonomy-Driven Synthetic Dataset for Retrieval Effectiveness and Fairness Evaluation in Hiring}

\author{Warren Jouanneau}
\correspondingauthor
\email{warren.jouanneau@malt.com}
\orcid{0000-0003-4973-2416}

\author{Morgane Hoffmann}

\author{Emma Jouffroy}
\orcid{0009-0003-2930-0330} 

\author{Marc Palyart}
\orcid{0000-0002-8808-0492}

\affiliation{%
  \institution{Malt}
  \city{Paris}
  \country{France}
}

\begin{abstract}
While machine learning hiring tools are ubiquitous, the need for privacy regulations and commercial sensitivity makes public benchmarking legally complex. We address this challenge by introducing a taxonomy-driven, fully synthetic HR dataset and benchmark to simultaneously evaluate retrieval effectiveness and algorithmic fairness. Grounded in 660K real freelancer profiles, our multilingual corpus features candidate profiles, project briefs, and counterfactual demographic twins. Auditing state-of-the-art embedding models reveals performance divergence from general MTEB scores, alongside fairness disparities in grammatically gendered languages. This demonstrates that general benchmarks are insufficient for HR, requiring tailored evaluations to identify Pareto-optimal models that balance semantic accuracy and fairness.\end{abstract}

\ccsdesc[500]{Information systems~Information retrieval}
\ccsdesc[500]{Computing methodologies~Natural language processing}
\ccsdesc[300]{Social and professional topics~Computing / technology policy}
\ccsdesc[100]{General and reference~Evaluation}

\keywords{Benchmark, Large Language Models, Retrieval, Hiring, Person-job Fit, Fairness}

\maketitle

\section{Introduction}

\input{01_Intro}

\section{Related Work}

\input{02_SOTA}

\section{Dataset Generation}
\input{03_Approach}

\section{Dataset Validation}
\label{sec:ds-validation}
\input{05_DS_Validation}

\section{Benchmark architecture and software implementation}
This section details the structural composition of our benchmark, its data distribution schemas, and the software environment engineered for execution. We present the dataset layout before outlining the evaluation framework architecture.

\input{06_Benchark}

\section{Results}
\input{07_results}

\section{Conclusion and Discussion}

We introduce a synthetic, taxonomy-grounded HR retrieval benchmark addresses privacy constraints while preserving much of its ecological validity. By evaluating a Train-on-Synthetic, Test-on-Real (TSTR) paradigm, we demonstrate that models trained exclusively on our synthetic data preserve substantial real-world utility. Specifically, TSTR captures 80\% of the retrieval performance gains achievable with proprietary data, yielding an 
NDCG@10 of 0.54 compared to 0.57 for the Train-on-Real, Test-on-Real (TRTR) configuration and 0.42 for the zero-shot baseline.

This dataset and benchmark are now publicly available in the MTEB format. While this framework effectively standardizes retrieval evaluation, our empirical audits reveal that traditional ranking metrics offer an incomplete picture of model viability: neither scaling parameters nor adopting modern training paradigms inherently neutralizes directional demographic bias. By structuring counterfactual fairness evaluation as a parallel MTEB task, we propose a dual-axis framework to systematically identify Pareto-optimal models that balance retrieval accuracy with demographic robustness.

\section{Limitations}
\label{sec:limitations}

While our synthetic approach reduces privacy risks and retains the downstream utility demonstrated above, inherent trade-offs remain in fidelity. Synthetic documents lack the natural structural diversity and formatting idiosyncrasies of real-world profiles, which subtly constrains the dataset's ability to perfectly mirror live distributions. Furthermore, relying on an LLM for relevance labeling across 1.6M interactions risks introducing algorithmic bias. Although models from our teacher's family have been shown to limit such factors~\cite{Hoffmann2026:llmbiashiring} and our synthetic scores achieved human-level agreement (Section~\ref{sec:synthetic-score-quality}), this manual validation was limited to 80 brief--profile pairs and excluded short queries.

Beyond data generation, the benchmark's scope is currently restricted to European languages and white-collar technology roles; therefore, evaluative findings may not directly transfer to other linguistic families or distinct labor markets (e.g., blue-collar trades). The static nature of the dataset also exposes it to temporal degradation as industry skills evolve, alongside a data contamination risk where future LLMs might ingest the benchmark during pre-training, thereby artificially inflating zero-shot evaluations.

Finally, offline evaluation metrics (
NDCG, 
MRR, 
IoU@k) serve as necessary but imperfect proxies that cannot fully capture live marketplace satisfaction. This limitation extends to our fairness framework, whose narrow construct (Section \ref{sec:counterfactual}) relies on a single human annotator for twin validation. Future work must extend this dual-axis evaluation to encompass other intertwined attributes, such as age and ethnicity, for instance through name-based signals as in correspondence studies \cite{bertrand2004emily} and embedding audits \cite{wilson2024gender}, enabling intersectional analyses.

\clearpage

\section*{Ethical Considerations Statement}

\paragraph{Source data and privacy.} The taxonomy was derived from 660K
freelancer profiles and 124K project records from a proprietary platform,
processed on the basis of the platform’s legitimate interest.
These records are not released, and only aggregate signals (job-title clusters,
skill frequencies, categorical attributes) informed generation. 
\paragraph{Annotators.} Relevance labels were validated by three members of the research team and twins by one of them, a researcher specializing in algorithmic fairness, as part of their paid working hours. They labelled only synthetic texts.
\paragraph{Scope and misuse.} Our fairness evaluation measures only how rankings
respond to switching grammatical gender markers between masculine and feminine
forms. It does not cover non-binary identities, other attributes, or indirect
proxies. Good results should not be read as evidence that a model is fair, and
the benchmark is not an audit or a conformity assessment of a hiring system
under the EU AI Act.
\paragraph{Inherited bias.} The taxonomy and the generating and scoring LLMs may
reproduce historical gender imbalances across occupations on the source platform.
Twin comparisons are not affected by imbalances in the base corpus, but relevance
labels may not be neutral.

\section*{Generative AI Usage Statement}
Large language models are a core component of the methodology of this work: Gemini 2.5 Pro, Gemini 2.5 Flash, and Gemini 3.5 Flash-Lite (Google) were used to generate the synthetic dataset, counterfactual twins, and relevance labels, as described in Section~4. Separately, during manuscript preparation, we used Claude Opus 5.5 (Anthropic) to suggest wording edits to some passages, and Gemini (Google) to provide feedback on a draft of the paper. All suggestions were reviewed by the authors, who accepted, revised, or discarded them. Neither tool was used to generate the paper's analyses, results, or citations, and the authors take full responsibility for its content.

\bibliographystyle{ACM-Reference-Format}
\bibliography{sample-base}

\appendix

\input{99_Sup_Mat}

\end{document}

%% file: 01_Intro.tex
Human resources (HR) increasingly relies on machine learning for candidate evaluation and job matching \cite{geyik2019fairness}, yet the methods used to validate these tools for bias are rarely fully transparent \cite{raghavan2020mitigating}. Furthermore, evaluating these models in real-world scenarios is often complicated by the commercial sensitivity and personally identifiable information (PII) inherent in professional data.

To bridge this evaluation gap, we present a taxonomy-driven, fully synthetic HR dataset and benchmark designed to simultaneously measure retrieval effectiveness and algorithmic fairness. Leveraging an empirical foundation of over 660K real freelancer profiles, we constructed a diverse, multilingual corpus of candidate records, project briefs, and counterfactual demographic twins, yielding 1.6M scored interactions.  

Auditing state-of-the-art text embedding models with this benchmark reveals several domain-specific insights. We find that model performance does not strictly correlate with parameter count or inference time. Furthermore, domain-specific evaluation matters:
for instance, Arctic Embed (v2, l) outperforms instruction-tuned models on project briefs by 0.07 NDCG@10 despite scoring 6--7 points lower on general MTEB multilingual benchmarks \cite{muennighoff2023:mteb}.
When evaluating grammatically gendered languages, we observe no systematic relationship between model scale and fairness: models of similar size can differ markedly in their robustness to gender perturbations. This work highlights that fairness and retrieval effectiveness are not necessarily conflicting objectives, and that tailored evaluations can identify Pareto-optimal models that balance both.

%% file: 02_SOTA.tex
Existing open-source benchmarks and datasets, such as Djinni \cite{Drushchak2024:DjinniDataset}, Xing \cite{Pacuk2016:RecSysChallengeHR}, CareerBuilder \cite{careerbuilder2012:kaggle}, and PJB \cite{Wang2026:PJBBench}, provide foundational HR data. However, they frequently lack multi-sided market visibility, suffer from label sparsity, or omit fairness evaluations entirely. Conversely, production-grade recommender systems rely on proprietary enterprise datasets (e.g., LinkedIn) that restrict standardized academic benchmarking. Furthermore, privacy regulations (\textit{e.g.}, the General Data Protection Regulation and the EU AI Act) rightfully safeguard personally identifiable information (PII) by restricting the distribution of real resumes. As highlighted by \cite{otani2025:NLPinHR}, these necessary protections create a systemic "missing data" challenge for HR research, prompting efforts like FairJob \cite{Vladimirova2024:FairJob} to rely on behavioral proxies for bias evaluation. Beyond the HR domain, generalized NLP fairness benchmarks like WinoBias \cite{Zhaoetal2018:WinoBias} and CrowS-Pairs \cite{Nangiaetal2020:CrowS-Pairs} isolate social biases but struggle to capture the complex semantic matching dynamics required in professional recruitment.

In the broader information retrieval literature, fairness in ranking has been formalized through constraints on top-$k$ representation \cite{zehlike2017fa} and group exposure \cite{singh2018fairness}, with fairness-aware re-ranking systems actively deployed in production talent search \cite{geyik2019fairness}. Yet, the specific methodologies used to validate and debias commercial hiring tools remain largely opaque, and vendors' fairness claims are often vague \cite{raghavan2020mitigating}. In economics, hiring discrimination is traditionally measured through correspondence studies that submit fictitious resumes differing only in a demographic signal, such as an applicant's name \cite{bertrand2004emily}. Our counterfactual twins transpose this design to evaluate retrieval models, drawing upon the framework of counterfactual fairness \cite{kusner2017counterfactual}. Closest to our work, \cite{wilson2024gender} audited embedding-based resume retrieval by introducing race- and gender-associated names to English resumes, revealing significant disparities. However, their evaluation did not assess retrieval effectiveness against graded relevance labels. Furthermore, in languages with grammatical gender, morphological marking shapes the representations of even inanimate nouns \cite{gonen2019does}, and the masculine and feminine forms of occupations are often encoded asymmetrically along the semantic gender direction \cite{zhou2019examining}. This linguistic complexity motivates our multilingual evaluation framework.

Modern candidate ranking relies heavily on bi-encoders and cross-encoders, which require high-fidelity matching signals. Zero-shot evaluation for these architectures is typically standardized via generalized benchmarks such as BEIR \cite{thakur2021:beir} and MTEB \cite{muennighoff2023:mteb}, which measure asymmetric text matching across diverse domains. To mitigate label sparsity and the high costs of manual annotation, researchers increasingly leverage Large Language Models (LLMs) to synthesize end-to-end evaluation datasets \cite{Rahmani2024:SyntheticRetrievalDataset}. This synthetic generation paradigm is now emerging within the HR domain to create resumes and job descriptions \cite{Yu2025:confitv2}, providing a promising avenue to bypass privacy constraints while retaining semantic complexity.

%% file: 03_Approach.tex
\begin{figure}
    \centering
    \includegraphics[width=1.\linewidth,trim={0 0 0 0},clip]{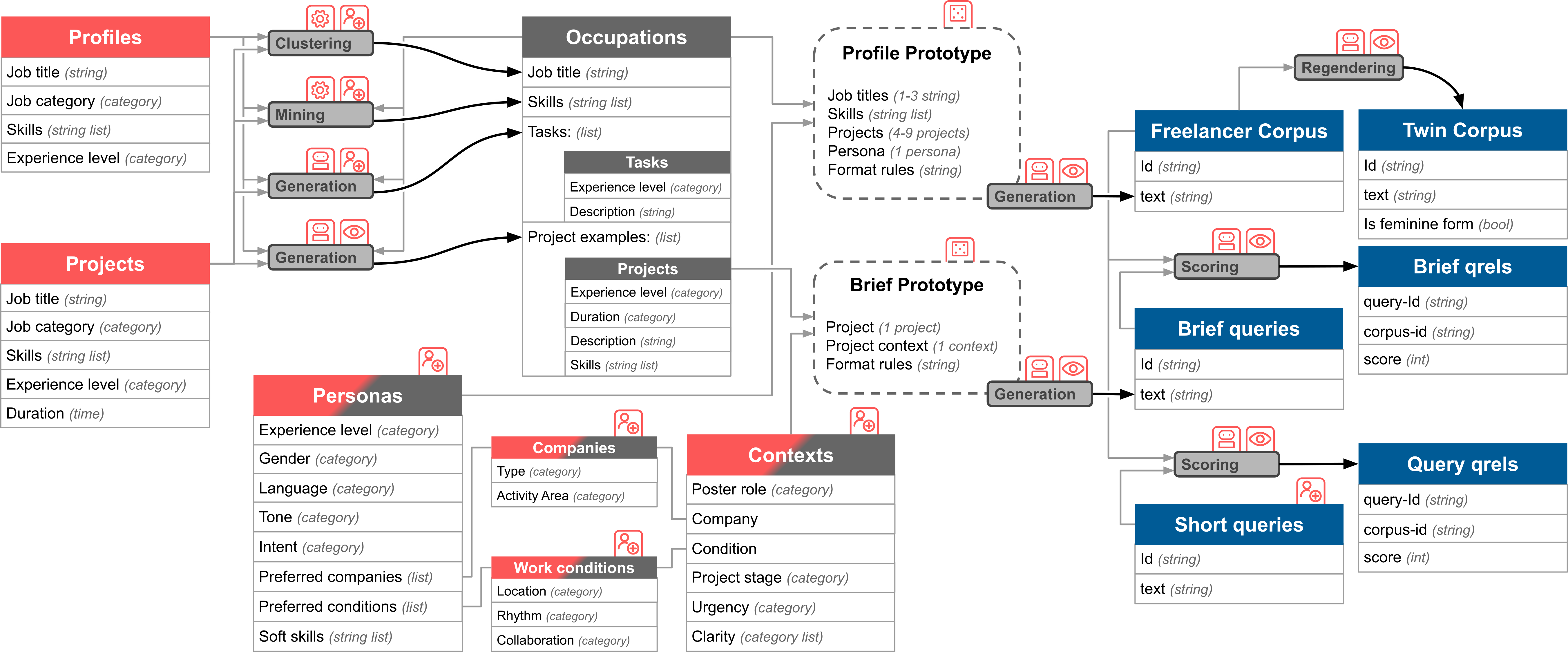}
    \caption{The synthetic dataset generation pipeline. Proprietary platform data (red) is abstracted into coarse-grained taxonomies and contexts (grey), which guide the LLM-driven synthesis and scoring of the final benchmark corpora (blue).}
    \Description{A left-to-right flowchart divided into three color-coded stages. On the left, red blocks representing proprietary platform data feed into a central grey section representing abstraction into taxonomies and contexts. Arrows from the grey section point to the right, where blue blocks depict the final LLM-driven synthesis and scoring of the benchmark corpora.}
    \label{fig:dataset-creation}
\end{figure}

To capture real-world labor dynamics while limiting the privacy risks of raw data, we synthesize an HR benchmark using models from the Gemini family~\cite{team2023gemini}, selected for each stage according to task complexity and computational cost: a larger model for taxonomy enrichment and document generation, and lighter models for large-scale relevance scoring and counterfactual twin generation.
Our four-stage pipeline (Figure \ref{fig:dataset-creation}) comprises: taxonomy extraction, document generation, synthetic interaction scoring, and counterfactual twin generation.

A core challenge in this process is maintaining \textit{ecological validity}, \textit{i.e.} ensuring offline evaluations faithfully reflect real-world performance. We achieve this by grounding our pipeline in macro-level empirical distributions. This abstraction allows us to capture authentic market signals, such as latent relationships between job titles, skills, and experience levels, without exposing proprietary business logic or leaking fine-grained PII. Structural fidelity and downstream utility analyses are detailed in Section \ref{sec:ds-validation}.

\subsection{Coarse-Grained Occupation Taxonomy}\label{sec:taxonomy}

Our dataset generation pipeline is grounded in a structured, coarse-grained occupational taxonomy (\textit{cf.} Figure~\ref{fig:dataset-creation}). Analyzing 660K anonymized freelancer profiles and 124K historical project records from a proprietary platform, we extracted and grouped job titles using text edit distance clustering. Subsequent human-in-the-loop curation consolidated these into 134 broad, distinct occupational clusters.

\begin{figure}
\centering
    \begin{tcolorbox}[academicexample]
    \textbf{Experience Level:} \texttt{INTERMEDIATE} \\
    \textbf{Duration:} \texttt{SHORT} \\[0.5em]
    \textbf{Description:} Develop a complete 5-page showcase website from a provided Figma design. You will use a premium theme and the Elementor page builder to create the pages, ensuring the final site is fully responsive and matches the design perfectly. \\[0.5em]
    \textbf{Skills:} Showcase website, Elementor, Webdesign, Responsive design, HTML5, CSS3 \\[0.5em]
    \textbf{Mobilized Tasks:}
    \begin{itemize}[nosep]
        \item Develop a complete showcase website from a provided design.
        \item Implement a fully responsive design from scratch using HTML5 and advanced CSS3 media queries.
        \item Convert a static web design into a functional WordPress theme or Elementor template.
    \end{itemize}
    \end{tcolorbox}
    \caption{\footnotesize WordPress developer project example.}
    \Description{A text box displaying an example project brief for a WordPress developer. It details the required experience level as intermediate, the duration as short, a text description for building a 5-page responsive website using Elementor, a list of six technical skills including HTML5 and CSS3, and a bulleted list of three specific mobilized web development tasks.}
\label{fig:occupation-project-example}
\end{figure}

For each occupation, we mined term frequencies from the source data to isolate statistically significant associated skills. To complete the taxonomy and abstract away any proprietary or fine-grained business logic, we used an LLM (Gemini 2.5 Pro \cite{Comanici2025:gemini2.5}) to generate realistic operational tasks and project examples tailored to each occupation and experience level (\texttt{ENTRY}, \texttt{INTERMEDIATE}, \texttt{EXPERT}, \texttt{EXPERT PLUS}). Tasks and projects are strictly differentiated: tasks define atomic execution capabilities, whereas projects aggregate multiple tasks over a specified duration: \texttt{SHORT} ($< 1$ month), \texttt{MEDIUM} ($1\text{--}6$ months), or \texttt{LONG} ($> 6$ months) (cf.~Figure~\ref{fig:occupation-project-example}).

To mirror realistic market dynamics, we enriched the taxonomy with Freelancer Personas and Project Contexts metadata. These signals govern both document \textit{content} and structural \textit{form}:

\noindent\textbf{Freelancer Personas:} 
The content of a freelancer profile is guided by core attributes including experience level, professional intent (\textit{e.g.}, skill Acquisition, portfolio Building, \dots), preferred company types (\textit{e.g.}, large luxury enterprise, early stage legal startup, \dots), desired work conditions (\textit{e.g.}, onsite squad Member, remote leader, \dots), and core soft skills. Stylistic form is driven by demographic attributes, language proficiencies, and communication tone (\textit{e.g.}, pedagogical, craft-focused, \dots).

\noindent\textbf{Project Contexts:} Project descriptions reflect company characteristics, project development stage (\textit{e.g.}, maintenance, consulting, \dots), and specification clarity. Phrasing is modulated by the poster's organizational role (\textit{e.g.}, Procurement, Recruiter, Tech Lead, \dots) and contract urgency (\textit{e.g.}, high-Urgency, standard, \dots).

By decoupling generation into macro-occupations and multi-dimensional persona attributes, the benchmark preserves high-fidelity, real-world labor configurations while avoiding the release of personal data.

\subsection{Taxonomy-Grounded Profile and Brief Generation}

To map our discrete taxonomy to natural language, we employ constrained LLM generation (Gemini 2.5 Pro \cite{Comanici2025:gemini2.5}) where taxonomic vectors serve as deterministic anchors. Freelancer profiles are instantiated by sampling combinations of one to three occupations and personas. We enforce strict combinatorial rules across the sampling space to prevent invalid feature intersections ( \textit{e.g.} pairing an \texttt{ENTRY}-level candidate with a highly strategic communication tone).

For each profile, we sample a portfolio of 4--9 past projects based on the persona's experience level and professional intent. For instance, an \texttt{INTERMEDIATE}-level profile receives 1--3 \texttt{ENTRY} and 3--6 \texttt{INTERMEDIATE} projects. Project durations are bound to career intent (e.g., a \textit{Stability \& Integration} intent maps to 5\% \texttt{SHORT}, 25\% \texttt{MEDIUM}, and 70\% \texttt{LONG} projects). The LLM then reformulates these text blocks, tailoring them to the candidate's skills and environmental preferences to form a coherent career track.

To simulate linguistic noise in user-generated text, we modulate the output across three formatting axes: 
(i)~\textit{typographical variation}, selectively injecting grammatical errors and typos; 
(ii)~\textit{structural layout}, constraining paragraph counts, text block lengths, and technical expressiveness; and 
(iii)~\textit{decorative elements}, modulating emoticon frequency based on the assigned communication tone.

Client briefs, serving as the evaluation queries in the benchmark, are generated via a parallel, symmetrical pipeline. Sampling task clusters and project durations from the shared taxonomy allows the LLM to simulate enterprise project specifications. This symmetric grounding establishes a clear semantic bridge between queries and profiles, capturing the latent vocabulary alignment and multi-sided retrieval challenges of freelance marketplaces.

\subsection{Synthetic Relevance Scoring}

Evaluating retrieval effectiveness and algorithmic fairness requires ground-truth skill-fit labels for each freelancer-brief pair. Following the methodology of \cite{jouanneau2026:SMS}, we prompt Gemini 3.5 Flash-Lite \cite{google_deepmind_gemini_2026} to generate a relevance score alongside a reasoning justification. To align with standard Information Retrieval (IR) practices, scores map to a discrete $\{-2, -1, 0, 1, 2, 3\}$ scale ranging from \textit{Severe Mismatch} ($-2$) to \textit{Optimal Match} ($3$), anchoring the numerical outputs to explicit qualitative interpretations.

Profiles are evaluated independently in randomized batches of ten. We compute the full Cartesian product of the evaluation space, scoring every generated profile against every client brief and short query.

\subsection{Counterfactual Augmentation for Fairness Evaluation}
\label{sec:counterfactual}

To evaluate gender bias while mitigating the uncontrolled confounding variables typical of real-world data, we augment our dataset with counterfactual pairs. This design adapts the methodology of correspondence studies in hiring discrimination \cite{bertrand2004emily} to the computational framework of counterfactual fairness \cite{kusner2017counterfactual}. We employ a stratified sampling strategy across the 134 occupations and five languages represented in the dataset. For each sampled profile, we generate a counterfactual twin that differs exclusively in textual gender cues (\textit{e.g.}, pronouns, grammatical agreements, and gendered occupational titles), holding all professional qualifications, skills, and technical content strictly constant.

Consequently, the fairness construct we measure is intentionally precise: the sensitivity of a retrieval model to masculine versus feminine morphological markers in otherwise identical profiles, which we treat as a proxy for disparate treatment based on presented gender. We acknowledge that this formulation does not capture the full spectrum of social gender identity beyond a grammatical binary, nor does it account for indirect demographic signals such as applicant names, historical career gaps, or broader occupational stereotypes. However, within this explicitly defined scope, grammatical gender presentation remains the sole manipulated variable. This ensures that any observed ranking disparities between twins can be directly attributed to algorithmic sensitivity rather than latent differences in profile quality or semantic content (see Appendix \ref{annex:twin_creation} for further generation and validation details).

%% file: 05_DS_Validation.tex
As introduced in our methodology, maintaining ecological validity ensures that offline evaluations on our synthetic benchmark faithfully reflect real-world performance. This section details the quantitative experiments conducted to validate the dataset's structural fidelity, the integrity of its synthetic annotations, and its practical utility for downstream machine learning tasks.

\subsection{Ecological Fidelity}

\begin{table}[htbp]
    \centering
    \caption{Descriptive statistics of profile texts across the synthetic dataset, an equivalently sized sample from the live platform, and the foundational clusters used for generation.}
    \begin{tabular}{lccc}
    \toprule
    \textbf{Metric} & \textbf{Synthetic} & \textbf{Platform} & \textbf{Clusters} \\
    \midrule
    Type-Token Ratio       & 0.70 & 0.74 & 0.76 \\
    Vocabulary Size        & 22k  & 33k  & 24k  \\
    Tokens (Mean)          & 142  & 114  & 95   \\
    Paragraphs (Mean)      & 2.25 & 3.58 & 2.65 \\
    Emojis (Mean)          & 0.29 & 0.29 & 0.11 \\
    \bottomrule
    \end{tabular}
    \label{tab:db-stats-eco-val}
\end{table}

Comparing the synthetic dataset to its foundational clusters reveals the expected dynamics of narrative expansion. To transition from discrete taxonomic inputs (mean tokens = 95) to cohesive profiles, the LLM introduces necessary grammatical and syntactic structures, resulting in longer profiles (mean tokens = 142) and a naturally lower Type-Token Ratio (0.70 vs.\ 0.76). The injection of contextual personas successfully elevates the stylistic tone of the raw cluster data, matching the synthetic emoji distribution with the live platform baseline (0.29).

However, maintaining strict semantic adherence to the source taxonomy inherently limits variance. Because the generative pipeline is fundamentally grounded in the foundational clusters, the synthetic vocabulary size is necessarily bottlenecked by this source material (22k for the synthetic dataset vs.\ 24k for the clusters). While this demonstrates high fidelity to the target inputs, it simultaneously prevents the synthetic dataset from capturing the unconstrained lexical diversity of the live platform, which features a significantly broader vocabulary (33k) and a higher Type-Token Ratio (0.74). Furthermore, synthetic profiles exhibit greater structural uniformity than their real-world counterparts: although they are longer on average (142 vs.\ 114 tokens), the text is consolidated into fewer paragraphs (2.25 vs.\ 3.58). Together, these lexical and structural constraints illustrate the inherent ecological fidelity trade-offs of our synthetic approach, as discussed in Section ~\ref{sec:limitations}.

\subsection{Synthetic Score Quality}
\label{sec:synthetic-score-quality}

To validate the ground truth skill-fit generated for each interaction, three annotators independently scored 80 brief--profile pairs using the same guidelines provided to the generative model \cite{jouanneau2026:SMS}. We focused this manual audit exclusively on project briefs, as their semantic complexity and length demand much more nuanced relevance judgments compared to explicit, straightforward short queries.

We report four agreement metrics, each computed for a pair of raters (either two human annotators, or the model and one human annotator). \textit{Fuzzy agreement} corresponds to the share of brief–profile pairs scored within one level of each other, while \textit{relevant agreement} measures the share placed on the same side of the relevance threshold ($\geq 1$). Cohen's kappa ($\kappa$) measures chance-corrected exact agreement; we report its quadratically weighted variant, which penalizes disagreements by their squared distance, and its binarized variant, computed strictly on relevance labels ($\geq 1$).

\begin{table}[htbp]
    \centering
    \caption{Agreement on 80 brief--profile pairs among human annotators (mean and range over three annotator pairs) and between synthetic scores and annotators (mean and range over three annotators).}
    \begin{tabular}{lcccc}
        \toprule
        & \multicolumn{2}{c}{\textbf{Humans}} & \multicolumn{2}{c}{\textbf{Synthetic}} \\
        \cmidrule(lr){2-3} \cmidrule(lr){4-5}
        \textbf{Metric} & Mean & Range & Mean & Range \\
        \midrule
        \% fuzzy agreement ($\pm 1$) & 0.857 & 0.797--0.937 & 0.873 & 0.785--0.962 \\
        \% relevant agreement ($\geq 1$) & 0.806 & 0.747--0.899 & 0.848 & 0.747--0.924 \\
        Quadratic Weighted $\kappa$ & 0.737 & 0.676--0.856 & 0.787 & 0.687--0.860 \\
        Binarized $\kappa$ ($\geq 1$) & 0.593 & 0.487--0.756 & 0.660 & 0.484--0.799 \\
        \bottomrule
    \end{tabular}
    \label{tab:synthetic-score-quality}
\end{table}

As shown in Table~\ref{tab:synthetic-score-quality}, the synthetic scores agree with the annotators at least as well as the annotators agree with each other on every metric, with synthetic means consistently falling within the range of pairwise human agreement. Large disagreements are rare: raters differ by two or more levels on only 13--14\% of pairs, and weighted kappa indicates substantial agreement for both humans (0.737) and the model (0.787) \cite{landis1977measurement}. Binarized kappa is lower (0.593 and 0.660), as this metric counts any deviation across the relevance threshold as a full disagreement, whereas the weighted variant penalizes adjacent-level disagreements only lightly. Nonetheless, raters agree on binary relevance for 81--85\% of pairs. Agreement is lowest for pairs involving the same specific annotator, both among humans and with the model (Appendix~\ref{annex:pairwise-agreement}), pointing to individual human subjectivity rather than systematic model errors. Synthetic scores thus reliably reach human-level agreement, validating their use as ground-truth relevance labels for the benchmark.

\subsection{Ecological Utility}

\begin{table}[htbp]
    \centering
    \caption{Comparison of retrieval performance at cutoff 10 under Zero-Shot, Train-on-Real Test-on-Real (TRTR), and Train-on-Synthetic Test-on-Real (TSTR) settings.}
    \begin{tabular}{lccc}
    \toprule
    \textbf{Training Setting} & \textbf{NDCG} & \textbf{P} & \textbf{MRR} \\
    \midrule
    Zero-Shot Baseline          & 0.42  & 0.55  & 0.76  \\
    TRTR (Historical Data)      & 0.57  & 0.72  & 0.87  \\
    TSTR (Synthetic Data)       & 0.54  & 0.70  & 0.83  \\
    \bottomrule
    \end{tabular}
    \label{tab:db-stats-eco-util}
\end{table}

To quantitatively assess the ecological utility of our synthetic dataset, we conducted an auxiliary cross-finetuning experiment using the Arctic Embed (v2, l) model. To establish an empirical reference point, we compiled a proprietary historical training dataset equivalent in size to our synthetic corpus. Note that this historical dataset serves strictly as an experimental control for this utility verification and is not part of the public benchmark release. Both models were then evaluated on a shared holdout test set of real historical interactions.

The zero-shot baseline achieves an $\text{NDCG}@10$ of 0.42, a $\text{P}@10$ of 0.55, and an $\text{MRR}@10$ of 0.76. Establishing the empirical upper bound, the Train-on-Real, Test-on-Real (TRTR) setting improves performance across all metrics ($\text{NDCG}@10 = 0.57$, $\text{P}@10 = 0.72$, $\text{MRR}@10 = 0.87$). In our Train-on-Synthetic, Test-on-Real (TSTR) setting, where the model is finetuned exclusively on our generated benchmark data, it captures 80\% of the NDCG improvements achieved by the TRTR upper bound ($\text{NDCG}@10$ of 0.54). While a minor performance gap remains—expected given the TRTR setting's natural advantage of identical training and testing distributions—this experiment demonstrates that our synthetic benchmark acts as a useful proxy for downstream machine learning optimization without releasing real user data.

%% file: 06_Benchark.tex
\subsection{Dataset Presentation}

Our dataset is hosted on Hugging Face under the identifier \texttt{MaltCommunity/fair-freelancer-retrieval}\footnote{\url{https://huggingface.co/datasets/MaltCommunity/fair-freelancer-retrieval}} and distributed under a CC BY 4.0 license.
The repository contains a primary subset for ranking effectiveness alongside multiple evaluation subsets for counterfactual fairness audits. These subsets are partitioned by protected demographic attributes, currently evaluating presented gender (\texttt{gender}) across five languages (English (\texttt{en}), French (\texttt{fr}), Spanish (\texttt{es}), German (\texttt{de}), and Dutch (\texttt{nl})) and remain extendable to future attributes like age or nationality. Every subset complies with the standard MTEB retrieval format (\texttt{corpus}, \texttt{queries}, \texttt{qrels}). In total, the dataset comprises $4,019$ freelancer profiles, $669$ counterfactual twins, $200$ short search queries, $200$ project briefs, and $1,607,600$ annotated relevance scores, supporting evaluations of both keyword-based talent \textit{search} and contextual talent \textit{recommendation}.

\subsection{Evaluation Framework and Effectiveness Metrics}

We implement our infrastructure as a programmatic extension of the MTEB library, integrating with its core retrieval loops to extract model predictions for downstream counterfactual fairness processing. Formally, given a set of evaluation queries $\mathcal{Q}$ and a pool of candidate profiles $\mathcal{P}$, the model under review computes a continuous similarity score $s(q, p)$ for each pair $(q, p) \in \mathcal{Q} \times \mathcal{P}$. These scores are evaluated against the ground-truth relevance labels $rel(q, p) \in \{-2, -1, 0, 1, 2, 3\}$.

To quantify retrieval performance, we report the macro-average across the query set $\mathcal{Q}$ for three standard metrics evaluated at rank $k$. For metrics requiring binary relevance, we set the positive threshold at $rel(q, p) \geq 1$:

\paragraph{\textbf{Precision} ($\text{P}@k$):} Measures the fraction of retrieved profiles within the top $k$ results that are relevant to the query. In a retrieval context, it captures the immediate density of valid matches in the user's primary field of view. For a given query $q$, let $\mathbb{I}(\cdot)$ be an indicator function for binary relevance ($rel \geq 1$). Precision at $k$ is defined as:
\begin{equation}
    \text{P}@k = \frac{1}{k} \sum_{r=1}^{k} \mathbb{I}[rel(q, p_r) \geq 1]
\end{equation}
where $p_r$ is the candidate profile retrieved at rank $r$.

\paragraph{\textbf{Mean Reciprocal Rank} ($\text{MRR}@k$):} Evaluates the rank position of the \textit{first} relevant profile across all queries. This metric reflects the search effort required for a user to satisfy their information need, heavily penalizing retrieval systems that bury valid matches. It is defined as:
\begin{equation}
    \text{MRR}@k = \frac{1}{|\mathcal{Q}|} \sum_{q \in \mathcal{Q}} \frac{1}{\text{rank}_q^*}
\end{equation}
where $\text{rank}_q^*$ is the position of the highest-ranked relevant profile for query $q$ (with $\frac{1}{\text{rank}_q^*} = 0$ if no relevant profile appears within the top $k$).

\paragraph{\textbf{Normalized Discounted Cumulative Gain} ($\text{NDCG}@k$):} Assesses overall retrieval quality using the graded relevance scale. In accordance with standard \texttt{pytrec\_eval} implementation, negative relevance labels ($-2$ and $-1$) are zeroed out during the DCG computation, treating severe mismatches equivalently to neutral or unjudged documents. It heavily rewards algorithms that place optimal matches ($rel=3$) at the very top of the ranking, applying a logarithmic decay to discount the value of documents retrieved at lower ranks. For a query $q$, the Discounted Cumulative Gain (DCG) is:
\begin{equation}
    \text{DCG}@k = \sum_{r=1}^{k} \frac{2^{\max(0, rel(q, p_r))} - 1}{\log_2(r + 1)}
\end{equation}
The $\text{NDCG}@k$ is then computed as $\frac{\text{DCG}@k}{\text{IDCG}@k}$, where the Ideal DCG (IDCG) represents the maximum possible score obtained by perfectly sorting all available relevant profiles in descending order of their ground-truth grades.

\subsection{Fairness Evaluation Framework and Metrics}

To operationalize counterfactual fairness auditing, we structure the evaluation as a parallel MTEB task. For each candidate $i$ and query $q \in \mathcal{Q}$, we fix the competitive background by computing retrieval scores for all competing candidates $j \in \mathcal{P} \setminus \{i\}$ using their original profiles. We then independently rank the demographic variants $p_i^{g}$ and $p_i^{\bar{g}}$ of candidate $i$ within this unchanged environment, isolating the effect of group presentation. Formally, for a variant $G \in \{g, \bar{g}\}$, the rank of $p_i^{G}$ is defined as:
\begin{equation}
    \text{Rank}(p_i^{G}, q) = 1 + \sum_{j \in \mathcal{P} \setminus \{i\}} \mathbf{I}[s(q, p_j) > s(q, p_i^{G})].
\end{equation}
We define the signed rank difference between variants as:
\begin{equation}
    \Delta_{i,q} = \text{Rank}(p_i^{\bar{g}}, q) - \text{Rank}(p_i^{g}, q).
\end{equation}
Because lower ranks denote better positions, a positive $\Delta_{i,q}$ indicates a ranking advantage for group $g$.

To systematically evaluate treatment differences, we map potential fairness metrics across three evaluation axes and three levels of granularity (Table \ref{tab:fairness_metrics_granularity_axis}). To faithfully reflect real-world user interactions with discrete search engine results, we discard continuous, score-based evaluations. In alignment with established fair ranking literature, our evaluation prioritizes a candidate's discrete visibility within the top-$k$ results \cite{zehlike2017fa} and the overall exposure they receive \cite{singh2018fairness}, although we defer exposure-weighted variants of our rank metrics to future work. From the remaining discrete metrics, we construct a comprehensive triad designed to rigorously quantify disparities across all evaluation axes: $\text{IoU}@k$ to assess top-$k$ ranking stability, Absolute Rank Shift to measure the magnitude of ranking perturbations, and Rank-Biserial Correlation (RBC) to capture systematic directional bias.

\begin{table}[htbp]
\centering
\caption{Axis and granularity of Fairness metrics}
\label{tab:fairness_metrics_granularity_axis}
\begin{tabular}{lccc}
\toprule
\textbf{Axis} & \textbf{Scores} & \textbf{Ranks} & \textbf{Top-$k$} \\ 
\midrule
Stability     & Pearson $r$     & Kendall's $\tau$ & IoU@$k$  \\ 
Direction     & Cohen's $d$     & RBC              & OddRatio@$k$       \\ 
Magnitude     & Std MAD         & Abs. Shift       & Dropout@$k$  \\ 
\bottomrule
\end{tabular}
\end{table}

The formal mathematical definitions for our selected triad of metrics are as follows:

\paragraph{\textbf{Intersection over Union} $(\text{IoU}@k)$:} Measures the stability of top-$k$ visibility under perturbations. For each candidate-query pair $(i, q) \in \mathcal{D}$, we evaluate whether both variants of the same underlying profile appear within the top $k$ retrieved results\footnote{By convention, if no profile variant appears in the top $k$ across all pairs (yielding a denominator of zero), the metric is undefined.}. Letting $R(\cdot)$ denote $\text{Rank}(\cdot, q)$ for brevity, it is defined as:
    \begin{equation}
    \text{IoU@}k = \frac{
    \sum_{(i,q)\in\mathcal D} \mathbb{I}\Big[ R(p_i^{\bar{g}}) \leq k \land R(p_i^{g}) \leq k \Big]
    }{
    \sum_{(i,q)\in\mathcal D} \mathbb{I}\Big[ R(p_i^{\bar{g}}) \leq k \lor R(p_i^{g}) \leq k \Big]
    }
    \end{equation}
It ranges from 0 to 1. Values near 1 suggest demographic perturbations minimally impact top-ranked profiles, while values near 0 indicate strong instability. Intermediate values represent partial overlap.

\paragraph{\textbf{Mean Absolute Rank Shift}:} Captures the average magnitude of ranking perturbations independent of directionality. It is defined as:
\begin{equation}
\text{MeanRankShift} = \frac{1}{|\mathcal{D}|} \sum_{(i,q)\in\mathcal{D}} |\Delta_{i,q}|.
\end{equation}
Higher values signify that perturbations substantially alter profile ranking positions. Because Mean Absolute Rank Shift averages over all candidate–query pairs, it is dominated by shifts deep in the ranking, which have little practical effect; we therefore use it as a measure of overall sensitivity and rely on IoU@$k$ for effects on the visible top of the ranking.

\paragraph{\textbf{Rank-Biserial Correlation} (RBC):} Measures the direction of disparities across evaluated pairs to capture whether a specific group presentation is systematically favored. Let $r_{i,q}$ denote the magnitude rank of the absolute perturbation $|\Delta_{i,q}|$. The cumulative ranks for positive and negative differences are:
\begin{equation}
\label{eq:rbc}
    W^+ = \sum_{\Delta_{i,q} > 0} r_{i,q} \quad \text{and} \quad W^- = \sum_{\Delta_{i,q} < 0} r_{i,q}.
\end{equation}
The RBC is then defined as:
\begin{equation}
    \text{RBC} = \frac{W^+ - W^-}{W^+ + W^-}.
\end{equation}
RBC ranges from -1 to 1. Values near zero indicate the absence of systematic directional disparities. Positive values suggest perturbations tend to favor group $g$, while negative values indicate an advantage for group $\bar{g}$. Because pairs with $\Delta_{i,q} = 0$ do not enter Eq.\ref{eq:rbc} and RBC ignores the size of shifts, it should be read jointly with Rank Shift: an extreme RBC paired with a negligible Rank Shift indicates a consistent but practically insignificant direction.

%% file: 07_results.tex
We evaluate our benchmark across a diverse suite of state-of-the-art open-weight embedding models, selected to represent the primary architectural and training paradigms driving modern retrieval. We include the E5 family (small, base, large)~\cite{wang2022:e5} as a robust baseline for dense retrieval driven by weakly-supervised contrastive learning. To assess unified search capabilities, we include BGE-m3~\cite{chen2024:m3}, which integrates multi-lingual dense, sparse, and multi-vector retrieval into a single framework. We also evaluate the English-focused Arctic Embed series (s, m, l)~\cite{merrick2024:arctic}, optimized specifically for enterprise search effectiveness, along with its successor, Arctic Embed (v2, l)~\cite{Yu2024:arcticembed2}, to assess how these capabilities transfer into a highly optimized multilingual space. To test inference-scaling techniques in search, we incorporate Nomic Embed (v2)~\cite{Nussbaum2025:nomicembed}, leveraging a Mixture-of-Experts (MoE) architecture to increase retrieval capacity without proportional compute overhead. Finally, we include PPLX Embed (0.6B)~\cite{Eslami2026:pplxembed} and Qwen3 Embedding (0.6B)~\cite{li2026:qwen3} to represent the latest frontier of instruction-tuned, LLM-based embeddings, which allow for highly contextual, task-aware document ranking.

This diversity allows us to investigate how underlying architectures and training data influence trade-offs between representational capacity, inference speed, and algorithmic fairness. To ensure equitable efficiency comparisons, all inferences were executed on a single Intel Core i7-12700H CPU with a batch size of 32.

To quantify the statistical stability of our evaluation metrics, we compute 95\% confidence intervals using a non-parametric bootstrapping approach. Rather than relying on single point estimates, we first compute all retrieval and fairness metrics at the individual brief and short query level. We then independently draw 1,000 random samples with replacement from the brief set, and 1,000 random samples with replacement from the short query set. 

For each resampled subset, we recalculate the macro-averaged metrics to construct an empirical distribution of model performance. The 95\% confidence intervals are derived by extracting the $\alpha = 0.025$ and $1 - \alpha = 0.975$ percentiles from this distribution. This resampling process indicates wether our reported effectiveness scores and fairness disparities are stable to query-level variations and not artifacts of specific brief--profile pairings.

\subsection{Effectiveness on Ranking} 

\begin{figure*}
    \centering
    \includegraphics[width=1.\linewidth,trim={0 0 0 20pt}, clip]{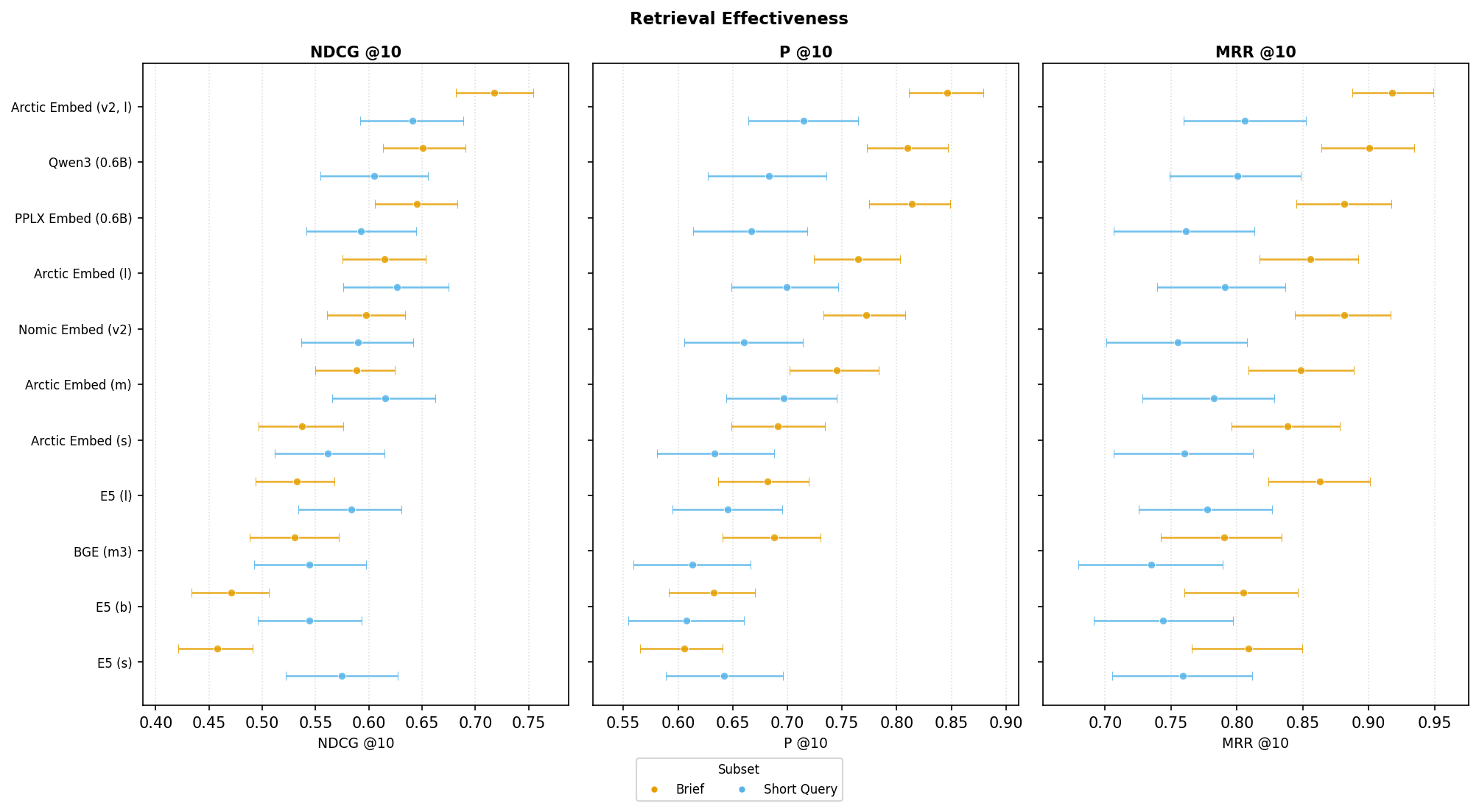}
    \caption{Retrieval effectiveness (NDCG@10, P@10, MRR@10) across embedding models for Brief and Short Query subsets.\\ \textit{Error bars represent 95\% bootstrap confidence intervals. Models are sorted descending by Brief NDCG@10.}}
    \Description{A horizontal dot-and-whisker plot displaying retrieval effectiveness across 11 embedding models[cite: 8]. The plot is divided into three panels for NDCG@10, P@10, and MRR@10[cite: 8]. Each model shows two error bars representing the 95\% confidence intervals: an orange dot for the Brief subset and a blue dot for the Short Query subset[cite: 8]. The models are ordered descending by Brief performance, with Arctic Embed (v2, l) showing the highest scores and E5 (s) the lowest[cite: 8].}
    \label{fig:ci-effectiveness}
\end{figure*}

\input{tables/retrieval-effectiveness}

As detailed in Table \ref{table:bench-effectiveness-results} and visualized with 95\% confidence intervals in Figure \ref{fig:ci-effectiveness}, ranking performance does not strictly correlate with parameter scale. For long briefs, Arctic Embed (v2, l) leads across all metrics ($\text{NDCG}@10 = 0.72$), while instruction-tuned models (Qwen3, PPLX Embed) trail at a clear, statistically distinct distance (e.g., $\text{NDCG}@10 = 0.65$) despite up to $7.7\times$ higher latencies. For short queries, Arctic Embed (v2, l) leads ($\text{NDCG}@10 = 0.64$) closely followed by Arctic Embed (l) (0.63), and Arctic Embed (m) actually outperforms heavier instruction-tuned models (0.62 vs. 0.59–0.60) at a fraction of their test time.

Efficiency gains are pronounced both across and within model generations: with just 33M parameters, Arctic Embed (s) matches the 278M-parameter E5 (base) on short queries and surpasses it on briefs. Within the Arctic family, Arctic Embed (m) trails Arctic Embed (l) by only 0.02 $\text{NDCG}@10$ on both tasks, utilizing roughly a third of its parameters and test time. Nomic Embed (v2) nearly matches Arctic Embed (l) on briefs at less than half its test time, but falls behind on short queries.

These gaps reflect the trade-offs between training data and architecture. The E5 family is limited by older contrastive data and standard architecture, whereas instruction-tuned LLMs capture complex semantics at the expense of high inference overhead. The Arctic family offers the strongest production trade-off, combining competitive effectiveness with low inference cost across model sizes. Arctic Embed (l) ranks among the best models on HR short queries despite a low general MTEB score (39.66). Its low multilingual MTEB score partly reflects its English-focused training; its strong results here suggest that HR retrieval may rely heavily on technical vocabulary shared across languages, such as skill and tool names. Similarly, Arctic Embed (v2, l) outperforms PPLX Embed and Qwen3 on briefs by 0.07 $\text{NDCG}@10$, despite scoring 6--7 points lower on MTEB. Together, these discrepancies suggest that generalized benchmarks are not sufficient for specialized recruitment retrieval. 

\subsection{Fairness Analysis}

\begin{figure*}
    \centering
    \includegraphics[width=1.\linewidth,trim={0 0 0 20pt}, clip]{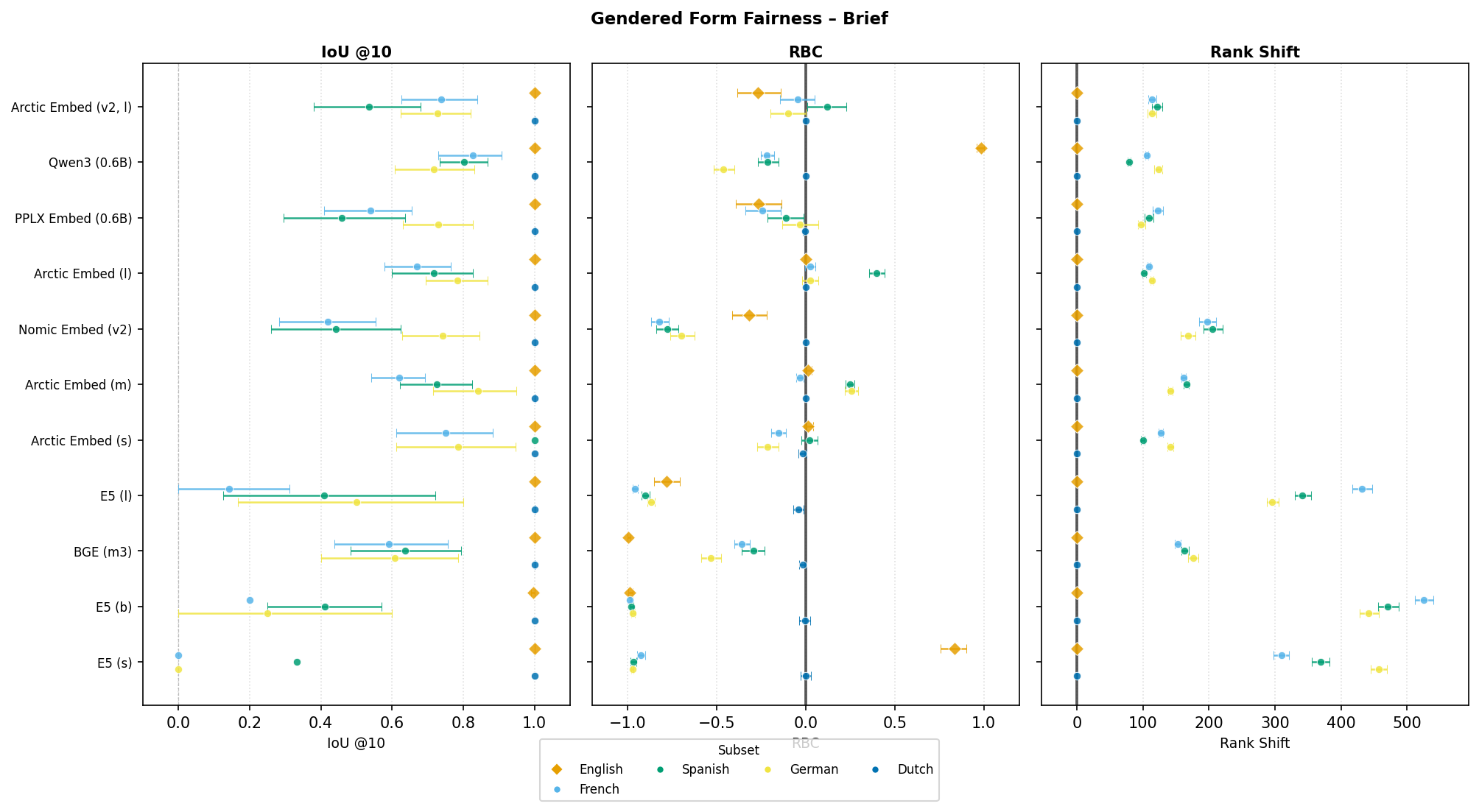}
    \caption{Gendered form fairness (IoU@10, RBC, Rank Shift) across embedding models for the Brief subset.\\ \textit{Error bars represent 95\% bootstrap confidence intervals. Models are sorted descending by Brief NDCG@10, identical to Figure \ref{fig:ci-effectiveness}. Grammatically gendered languages (French, Spanish, German) show lower ranking stability and larger rank shifts compared to English and Dutch. Directional bias (RBC) varies across models in all languages except Dutch; in English, extreme RBC values correspond to sub-position rank shifts and should not be read as substantial bias.}}
    \Description{A horizontal dot-and-whisker plot comparing gendered form fairness across 11 embedding models for the Brief subset[cite: 7]. The plot features three vertical panels: IoU@10, RBC, and Rank Shift, with 95\% confidence intervals displayed for each data point[cite: 7]. The y-axis lists models sorted by effectiveness. Five languages are plotted per model: English (orange diamonds), Dutch (dark blue dots), French (light blue dots), Spanish (green dots), and German (yellow dots)[cite: 7]. English and Dutch consistently align perfectly at 1.0 for IoU and near 0.0 for Rank Shift[cite: 7]. French, Spanish, and German display scattered data points with wide confidence intervals, indicating significant performance shifts and biases when gendered text is altered[cite: 7].}
    \label{fig:fairness-ci-brief}
\end{figure*}

\input{tables/gender/brief/retrieval-fairness}

Table \ref{table:bench-fairness-gender-brief-results} and the plotted confidence intervals in Figure \ref{fig:fairness-ci-brief} outline fairness metrics for project briefs (see Appendix \ref{annex:extended_fairness_analysis} for short queries). Given feminine ($g$) and masculine ($\bar{g}$) variants, a negative RBC indicates a systematic ranking advantage for masculine presentations.

Our results reveal pronounced cross-linguistic disparities in ranking stability. While English and Dutch rankings demonstrate near-perfect resilience across all evaluated models ($\text{IoU}@10 = 1.00$; mean rank shifts $< 1$), grammatically gendered languages (French, Spanish, German) show large ranking changes where it matters most, in the top 10: for E5 (base), IoU@10 falls to 0.20 in French, and E5 (small) drops to 0.00 in French and German. Rank shifts are also large on average (up to 525 positions for E5 (base) in French), although most of these occur far below the positions a recruiter would review.
The observed neutrality in Dutch (zero absolute rank shifts and $|\text{RBC}| \leq 0.04$ across all models) likely reflects the limited morphological gender marking and the contemporary shift toward gender-neutral occupational terminology in professional Dutch, which would leave twins close to their originals; we leave a per-language analysis of twin edits to future work.
In English, where twins differ only by a few pronouns, rank shifts remain below half a position for all models (at most 0.46). Some models nonetheless show a consistent direction when shifts do occur (e.g., $|\text{RBC}| \geq 0.98$ for E5 (base), BGE-m3, and Qwen3), but because RBC ignores both magnitude and ties, these values reflect a small number of minor shifts and have negligible practical impact on rankings.

Differences also emerge across model families. E5 models exhibit strong instability and negative directional bias (e.g., E5 (base) $\text{RBC} = -0.99$ in French and $-0.98$ in Spanish). 
This masculine advantage in grammatically gendered languages is consistent with prior findings that feminine occupation forms are encoded further from their English translations than masculine forms \cite{zhou2019examining}.
Moving away from contrastive learning does not automatically guarantee fairness. Arctic, Nomic, and BGE models improve overall stability over E5 but can retain strong directional bias (e.g., Nomic Embed (v2) $\text{RBC} = -0.82$ in French). Furthermore, ranking sensitivity alone is insufficient to characterize fairness: Arctic Embed (m) induces substantial rank shifts in French (162 positions) despite a near-neutral directional balance ($\text{RBC} = -0.03$). For the English-focused Arctic Embed models (s, m, l), stability in gendered languages should be interpreted with caution, as it may partly reflect limited sensitivity to their morphology rather than genuine robustness (e.g., Arctic Embed (s) reaches $\text{IoU}@10 = 1.00$ in Spanish). Ultimately, while scale alone does not guarantee fairness (Appendix \ref{annex:extended_fairness_analysis}), Pareto-optimal models like Arctic Embed (v2, l) balance high retrieval effectiveness with demographic robustness: on briefs, it combines the best $\text{NDCG}@10$ ($0.72$) with limited directional bias across gendered languages ($|\text{RBC}| \leq 0.12$; rank shifts of 113--121), showing these objectives are not always mutually exclusive.

%% file: tables/retrieval-effectiveness.tex
\begin{table*}[ht]
\caption{{Benchmark retrieval effectiveness results.\\\footnotesize\textit{ \colorbox[HTML]{DBDBDB}{\textbf{Best}} and \colorbox[HTML]{F0F0F0}{\textbf{Second Best}}values are highlighted per column.}}}
\centering
\small
\renewcommand{\arraystretch}{0.85}
\begin{tabularx}{\textwidth}{*{3}{l}*{3}{>{\centering\arraybackslash}X}*{3}{>{\centering\arraybackslash}X}*{1}{>{\centering\arraybackslash}X}}
\toprule
\multicolumn{3}{l}{\bfseries Model} & \multicolumn{3}{c}{\bfseries Brief (@10)} & \multicolumn{3}{c}{\bfseries Short Query (@10)} & {\bfseries MTEB}\\
 \cmidrule(lr){1-3} \cmidrule(lr){4-6} \cmidrule(lr){7-9}
Name (version) & Params & Test time & NDCG & P & MRR & NDCG & P & MRR & Score\\
\midrule
Arctic Embed (s) & 33M & 6m & 0.54 & 0.69 & 0.84 & 0.56 & 0.63 & 0.76 & 39.84 \\
E5 (small) & 118M & 6m & 0.46 & 0.61 & 0.81 & 0.57 & 0.64 & 0.76 & 36.25 \\
Arctic Embed (m) & 109M & 17m & 0.59 & 0.74 & 0.85 & 0.62 & \bfseries {\cellcolor[HTML]{F0F0F0}} 0.70 & 0.78 & 39.33 \\
E5 (base) & 278M & 18m & 0.47 & 0.63 & 0.80 & 0.54 & 0.61 & 0.74 & 37.69 \\
Nomic Embed (v2) & 475M & 25m & 0.60 & 0.77 & 0.88 & 0.59 & 0.66 & 0.75 & - \\
E5 (large) & 560M & 60m & 0.53 & 0.68 & 0.86 & 0.58 & 0.65 & 0.77 & 38.50 \\
Arctic Embed (l) & 335M & 61m & 0.61 & 0.76 & 0.86 & \bfseries {\cellcolor[HTML]{F0F0F0}} 0.63 & \bfseries {\cellcolor[HTML]{F0F0F0}} 0.70 & \bfseries {\cellcolor[HTML]{F0F0F0}}  0.79 & 39.66 \\
Arctic Embed (v2, l) & 568M & 107m & \bfseries {\cellcolor[HTML]{DBDBDB}} 0.72 & \bfseries {\cellcolor[HTML]{DBDBDB}} 0.85 & \bfseries {\cellcolor[HTML]{DBDBDB}} 0.92 & \bfseries {\cellcolor[HTML]{DBDBDB}} 0.64 & \bfseries {\cellcolor[HTML]{DBDBDB}} 0.71 & \bfseries {\cellcolor[HTML]{DBDBDB}} 0.80 & 58.36 \\
BGE (m3) & 568M & 108m & 0.53 & 0.69 & 0.79 & 0.54 & 0.61 & 0.73 & 54.59 \\
PPLX Embed (0.6B) & 596M & 207m & \bfseries {\cellcolor[HTML]{F0F0F0}}  0.65 & \bfseries {\cellcolor[HTML]{F0F0F0}} 0.81 & 0.88 & 0.59 & 0.67 & 0.76 & \bfseries {\cellcolor[HTML]{DBDBDB}} 65.41 \\
Qwen3 (0.6B) & 596M & 768m & \bfseries {\cellcolor[HTML]{F0F0F0}} 0.65 & \bfseries {\cellcolor[HTML]{F0F0F0}}  0.81 & \bfseries {\cellcolor[HTML]{F0F0F0}} 0.90 & 0.60 & 0.68 & \bfseries {\cellcolor[HTML]{DBDBDB}} 0.80 & \bfseries {\cellcolor[HTML]{F0F0F0}} 64.65 \\
\bottomrule
\end{tabularx}

\label{table:bench-effectiveness-results}
\end{table*}

%% file: tables/gender/brief/retrieval-fairness.tex
\begin{table*}[ht]
\caption{{Gendered form retrieval fairness results on Brief.\\\footnotesize\textit{ \colorbox[HTML]{DBDBDB}{\textbf{Best}} and \colorbox[HTML]{F0F0F0}{\textbf{Second Best}} values are highlighted per column.}}}
\centering\footnotesize\setlength{\tabcolsep}{1pt}
\begin{tabularx}{\textwidth}{*{1}{l}*{3}{>{\centering\arraybackslash}X}*{3}{>{\centering\arraybackslash}X}*{3}{>{\centering\arraybackslash}X}*{3}{>{\centering\arraybackslash}X}*{3}{>{\centering\arraybackslash}X}}
\toprule
\bfseries Model & \multicolumn{3}{c}{\bfseries English} & \multicolumn{3}{c}{\bfseries French} & \multicolumn{3}{c}{\bfseries Spanish} & \multicolumn{3}{c}{\bfseries German} & \multicolumn{3}{c}{\bfseries Dutch} \\
 \cmidrule(lr){1-1} \cmidrule(lr){2-4} \cmidrule(lr){5-7} \cmidrule(lr){8-10} \cmidrule(lr){11-13} \cmidrule(lr){14-16}
Name (version) & IoU @10 & RBC & Rank Shift & IoU @10 & RBC & Rank Shift & IoU @10 & RBC & Rank Shift & IoU @10 & RBC & Rank Shift & IoU @10 & RBC & Rank Shift \\
\midrule
Arctic Embed (s) & \bfseries {\cellcolor[HTML]{DBDBDB}} 1.00 & \bfseries {\cellcolor[HTML]{F0F0F0}} 0.01 & \bfseries {\cellcolor[HTML]{DBDBDB}} 0.00 & \bfseries {\cellcolor[HTML]{F0F0F0}} 0.75 & -0.15 & 127 & \bfseries {\cellcolor[HTML]{DBDBDB}} 1.00 & \bfseries {\cellcolor[HTML]{DBDBDB}} 0.02 & \bfseries {\cellcolor[HTML]{F0F0F0}} 100 & \bfseries {\cellcolor[HTML]{F0F0F0}} 0.79 & -0.21 & 141 & \bfseries {\cellcolor[HTML]{DBDBDB}} 1.00 & \bfseries {\cellcolor[HTML]{F0F0F0}} -0.01 & \bfseries {\cellcolor[HTML]{DBDBDB}} 0.00 \\
E5 (s) & \bfseries {\cellcolor[HTML]{DBDBDB}} 1.00 & 0.83 & 0.10 & 0.00 & -0.93 & 310 & 0.33 & -0.97 & 369 & 0.00 & -0.97 & 457 & \bfseries {\cellcolor[HTML]{DBDBDB}} 1.00 & \bfseries {\cellcolor[HTML]{DBDBDB}} 0.00 & \bfseries {\cellcolor[HTML]{DBDBDB}} 0.00 \\
Arctic Embed (m) & \bfseries {\cellcolor[HTML]{DBDBDB}} 1.00 & \bfseries {\cellcolor[HTML]{F0F0F0}} 0.01 & \bfseries {\cellcolor[HTML]{DBDBDB}} 0.00 & 0.62 & \bfseries {\cellcolor[HTML]{DBDBDB}} -0.03 & 162 & 0.73 & 0.25 & 166 & \bfseries {\cellcolor[HTML]{DBDBDB}} 0.84 & 0.26 & 142 & \bfseries {\cellcolor[HTML]{DBDBDB}} 1.00 & \bfseries {\cellcolor[HTML]{DBDBDB}} 0.00 & \bfseries {\cellcolor[HTML]{DBDBDB}} 0.00 \\
E5 (b) & \bfseries {\cellcolor[HTML]{DBDBDB}} 1.00 & -0.99 & 0.41 & 0.20 & -0.99 & 525 & 0.41 & -0.98 & 471 & 0.25 & -0.97 & 442 & \bfseries {\cellcolor[HTML]{DBDBDB}} 1.00 & \bfseries {\cellcolor[HTML]{F0F0F0}} -0.01 & \bfseries {\cellcolor[HTML]{DBDBDB}} 0.00 \\
Nomic Embed (v2) & \bfseries {\cellcolor[HTML]{DBDBDB}} 1.00 & -0.32 & \bfseries {\cellcolor[HTML]{F0F0F0}} 0.01 & 0.42 & -0.82 & 198 & 0.44 & -0.78 & 205 & 0.74 & -0.70 & 168 & \bfseries {\cellcolor[HTML]{DBDBDB}} 1.00 & \bfseries {\cellcolor[HTML]{DBDBDB}} 0.00 & \bfseries {\cellcolor[HTML]{DBDBDB}} 0.00 \\
E5 (l) & \bfseries {\cellcolor[HTML]{DBDBDB}} 1.00 & -0.78 & 0.05 & 0.14 & -0.96 & 431 & 0.41 & -0.90 & 341 & 0.50 & -0.87 & 296 & \bfseries {\cellcolor[HTML]{DBDBDB}} 1.00 & -0.04 & \bfseries {\cellcolor[HTML]{DBDBDB}} 0.00 \\
Arctic Embed (l) & \bfseries {\cellcolor[HTML]{DBDBDB}} 1.00 & \bfseries {\cellcolor[HTML]{DBDBDB}} 0.00 & \bfseries {\cellcolor[HTML]{DBDBDB}} 0.00 & 0.67 & \bfseries {\cellcolor[HTML]{DBDBDB}} 0.03 & \bfseries {\cellcolor[HTML]{F0F0F0}} 109 & 0.72 & 0.40 & 102 & 0.78 & \bfseries {\cellcolor[HTML]{DBDBDB}} 0.03 & \bfseries {\cellcolor[HTML]{F0F0F0}} 113 & \bfseries {\cellcolor[HTML]{DBDBDB}} 1.00 & \bfseries {\cellcolor[HTML]{DBDBDB}} 0.00 & \bfseries {\cellcolor[HTML]{DBDBDB}} 0.00 \\
Arctic Embed (v2, l) & \bfseries {\cellcolor[HTML]{DBDBDB}} 1.00 & -0.27 & 0.04 & 0.74 & \bfseries {\cellcolor[HTML]{F0F0F0}} -0.04 & 114 & 0.53 & 0.12 & 121 & 0.73 & -0.10 & \bfseries {\cellcolor[HTML]{F0F0F0}} 113 & \bfseries {\cellcolor[HTML]{DBDBDB}} 1.00 & \bfseries {\cellcolor[HTML]{DBDBDB}} 0.00 & \bfseries {\cellcolor[HTML]{DBDBDB}} 0.00 \\
BGE (m3) & \bfseries {\cellcolor[HTML]{DBDBDB}} 1.00 & -0.99 & 0.17 & 0.59 & -0.36 & 152 & 0.64 & -0.29 & 163 & 0.61 & -0.53 & 176 & \bfseries {\cellcolor[HTML]{DBDBDB}} 1.00 & \bfseries {\cellcolor[HTML]{F0F0F0}} -0.01 & \bfseries {\cellcolor[HTML]{DBDBDB}} 0.00 \\
PPLX Embed (0.6B) & \bfseries {\cellcolor[HTML]{DBDBDB}} 1.00 & -0.27 & 0.17 & 0.54 & -0.24 & 122 & 0.46 & \bfseries {\cellcolor[HTML]{F0F0F0}} -0.11 & 109 & 0.73 & \bfseries {\cellcolor[HTML]{DBDBDB}} -0.03 & \bfseries {\cellcolor[HTML]{DBDBDB}} 97 & \bfseries {\cellcolor[HTML]{DBDBDB}} 1.00 & \bfseries {\cellcolor[HTML]{F0F0F0}} -0.01 & \bfseries {\cellcolor[HTML]{DBDBDB}} 0.00 \\
Qwen3 (0.6B) & \bfseries {\cellcolor[HTML]{DBDBDB}} 1.00 & 0.98 & 0.46 & \bfseries {\cellcolor[HTML]{DBDBDB}} 0.83 & -0.22 & \bfseries {\cellcolor[HTML]{DBDBDB}} 106 & \bfseries {\cellcolor[HTML]{F0F0F0}} 0.80 & -0.21 & \bfseries {\cellcolor[HTML]{DBDBDB}} 80 & 0.72 & -0.46 & 123 & \bfseries {\cellcolor[HTML]{DBDBDB}} 1.00 & \bfseries {\cellcolor[HTML]{DBDBDB}} 0.00 & \bfseries {\cellcolor[HTML]{DBDBDB}} 0.00 \\
\bottomrule
\end{tabularx}

\label{table:bench-fairness-gender-brief-results}
\end{table*}

%% file: 99_Sup_Mat.tex
\newpage

\section{Additional details on counterfactual augmentation}
\label{annex:twin_creation}

This section provides the technical specifications and concrete examples of profile modifications and discusses the methodological challenges involved in extending it to other sensitive attributes. 

\paragraph{Targeted Fields and Counterfactual Perturbation Method.}
The transformation targets three distinct fields of the profiles likely to carry gender information: (i) the profile description, (ii) the job-title, and (iii) the titles, roles, and descriptions of work experiences. Within these fields, the perturbation consists in modifying explicit gendered linguistic markers, specifically pronouns, grammatical agreements, and gendered occupational noun, while holding all skills, profile length, and technical content constant. This method aims to isolate the gender variable.

\paragraph{Cross-Lingual Variance.}
The counterfactual transformation is inherently language-dependent. In highly gendered languages (French, Spanish, and German), gender presentation is deeply encoded through grammatical agreements, inflections, and occupational forms, requiring extensive rewriting. Grammatical gender is known to shape embedding representations well beyond the noun itself \cite{zhou2019examining}, so these rewrites are expected to move profiles in the embedding space.
In contrast, English profiles often remain nearly unchanged due to the more limited presence of explicit grammatical gender markers. This cross-lingual variance also serves as an internal robustness check for our proposed fairness metrics: minimal perturbations in English should theoretically yield negligible ranking variation compared to more gendered languages.

\paragraph{Generation and Validation Procedure.}
In practice, twin generation was performed using \textit{Gemini 2.5 Flash} with a prompt designed to minimize unwanted modifications. All generated twins were subsequently reviewed by a human annotator. The primary sources of potential errors include: 
(i) unintended modification of non-gendered technical terms;
(ii) hallucinated gender changes referring to third parties rather than the profile subject; and 
(iii) spurious formatting or structural alterations of the profile. 
Although such artifacts remained relatively rare in practice, manual validation was necessary to preserve non-demographic content across counterfactual pairs.

\paragraph{Concrete Transformation Example} 
To illustrate these linguistic adjustments consider the French profile description \textit{``Je suis un consultant expérimenté spécialisé dans les infrastructures cloud''} is transformed into \textit{``Je suis une consultante expérimentée spécialisée dans les infrastructures cloud''}. Similarly, in Spanish, the masculine occupational title \textit{``desarrollador senior''}  becomes \textit{``desarrolladora senior''}. In both cases, all technical skills, formatting, and project descriptions remain identical. \\

\paragraph{Methodological Challenges for Other Attributes.}
More generally, the proposed framework could be extended to additional demographic attributes beyond gender presentation, although constructing controlled counterfactual interventions becomes substantially more difficult for certain dimensions. In particular, some attributes are highly correlated with productive variables: age, for instance, is strongly correlated with years of professional experience, making it challenging to modify one without introducing unrealistic inconsistencies in the profile.
Ethnic origin also constitutes an important dimension for future fairness evaluations. While demographic perception is often entangled with explicit signals such as names or profile pictures, which are intentionally excluded from many retrieval systems to reduce discriminative bias, related information may still remain indirectly encoded in profile text through other observable profile attributes, such as foreign educational institutions or career trajectories.

\section{Extended Validation of Synthetic Scores}
\label{annex:pairwise-agreement}

Table \ref{tab:pairwise-agreement} provides a granular breakdown of the pairwise agreement metrics summarized previously in Table \ref{tab:synthetic-score-quality}. Within the human cohort, inter-rater reliability peaks between annotators A and B, whereas any pairing involving annotator C consistently yields the lowest consensus across all evaluated metrics. 

Crucially, the synthetic scores (M) follow the same pattern as human agreement. The model demonstrates high alignment with annotators A and B; in fact, the synthetic-to-human pairing of M--B outperforms the strongest human-to-human baseline (A--B) across all four metrics. Furthermore, the model's agreement with annotator C closely matches the lower consensus rates observed among the human raters themselves (e.g., quadratic weighted $\kappa = 0.687$ for M--C, compared to $0.679$ for A--C and $0.676$ for B--C). This pattern suggests that the lower aggregate agreement observed in Table \ref{tab:synthetic-score-quality} is an artifact of annotator C's idiosyncratic grading baseline, rather than a systematic failure or deviation in the synthetic scoring mechanism.

\begin{table}[htbp]
    \centering
    \footnotesize
    \setlength{\tabcolsep}{6pt}
    \caption{Pairwise agreement on 80 brief--profile interactions. The table compares inter-rater reliability among human annotators (A, B, C) against the alignment between the synthetic model scores (M) and each individual human.}
    \begin{tabular}{lcccccc}
        \toprule
        & \multicolumn{3}{c}{\textbf{Human--Human}} & \multicolumn{3}{c}{\textbf{Synthetic--Human}} \\
        \cmidrule(lr){2-4} \cmidrule(lr){5-7}
        \textbf{Metric} & A--B & A--C & B--C & M--A & M--B & M--C \\
        \midrule
        Fuzzy agreement ($\pm 1$) & 0.937 & 0.797 & 0.835 & 0.873 & 0.962 & 0.785 \\
        Relevant agreement ($\geq 1$) & 0.899 & 0.747 & 0.772 & 0.873 & 0.924 & 0.747 \\
        Quadratic Weighted $\kappa$ & 0.856 & 0.679 & 0.676 & 0.814 & 0.860 & 0.687 \\
        Binarized $\kappa$ ($\geq 1$) & 0.756 & 0.487 & 0.536 & 0.695 & 0.799 & 0.484 \\
        \bottomrule
    \end{tabular}
    \label{tab:pairwise-agreement}
\end{table}

\section{Extended Analysis}
\label{annex:extended_fairness_analysis}

\input{09_extended_results}

%% file: 09_extended_results.tex
Building upon the primary fairness metrics established in the main text, this appendix provides an extended analysis of how model scale, query length, and architectural choices interact with demographic bias, specifically focusing on the Short Query subset. 

\begin{figure*}[h]
    \centering
    \includegraphics[width=1.\linewidth,trim={0 0 0 20pt}, clip]{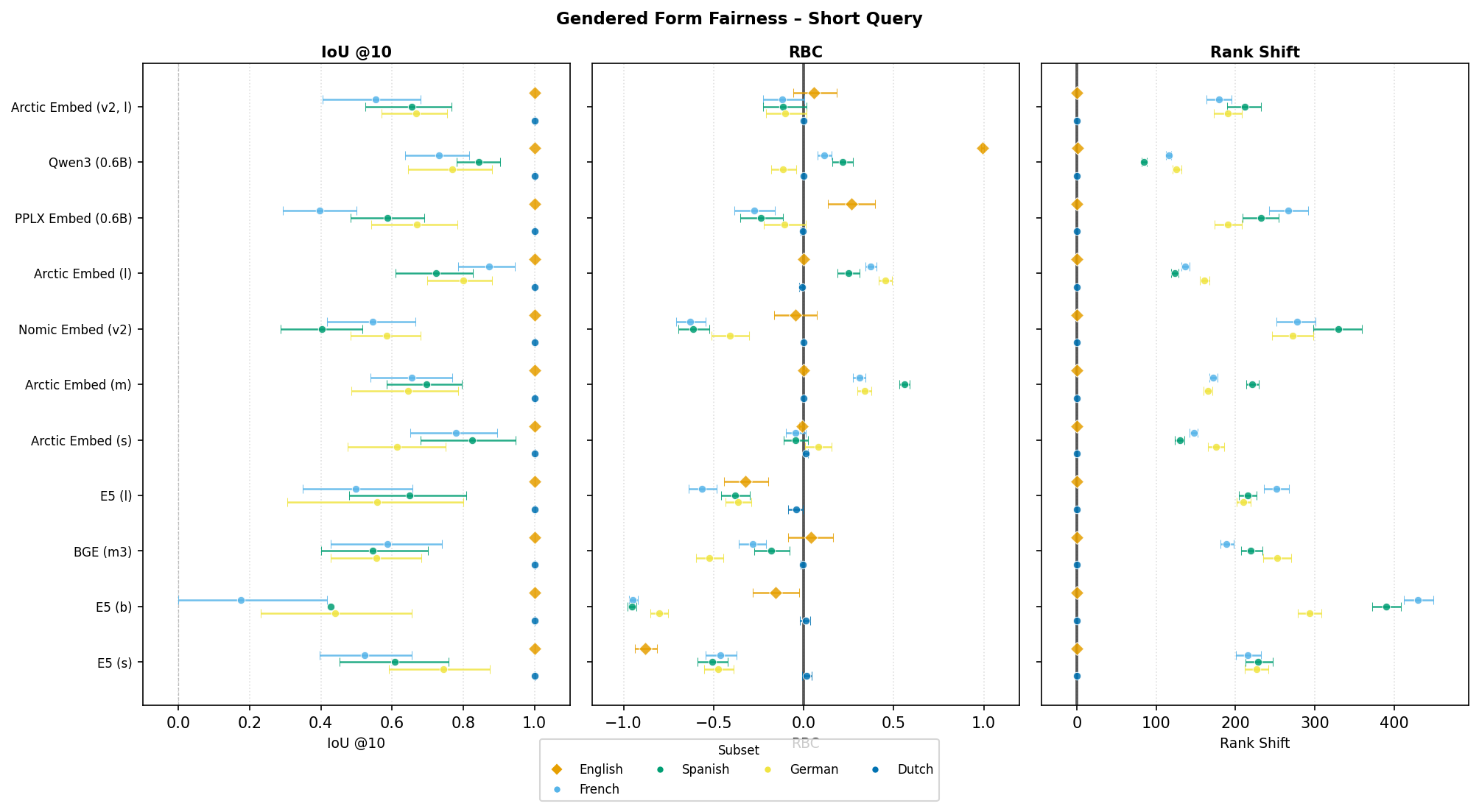}
    \caption{Gendered form fairness (IoU@10, RBC, Rank Shift) across embedding models for the Short Query subset.\\ \textit{Error bars represent 95\% bootstrap confidence intervals. Models are sorted descending by Brief NDCG@10, identical to the main text. Grammatically gendered languages (French, Spanish, German) exhibit lower ranking stability and substantial rank shifts, while English and Dutch remain fully stable (IoU@10 = 1.00). In English, some models show a consistent direction (e.g., RBC = 0.99 for Qwen3), but with rank shifts below one position.}}
    \Description{A horizontal dot-and-whisker plot comparing gendered form fairness across 11 embedding models for the Short Query subset[cite: 3]. The plot features three vertical panels: IoU@10, RBC, and Rank Shift, with 95\% confidence intervals displayed[cite: 3]. English (orange diamonds) and Dutch (dark blue dots) consistently plot exactly at the ideal baselines of 1.0 for IoU and 0.0 for RBC and Rank Shift, exhibiting zero variance[cite: 3]. In contrast, French, Spanish, and German show scattered data points with wide confidence intervals, indicating significant performance deviations and rank shifts[cite: 3].}
    \label{fig:fairness-ci-short-query}
\end{figure*}

\input{tables/gender/query/retrieval-fairness}

\textbf{Scale and Fairness.} A critical question is whether larger, highly-performant retrieval models are inherently fairer. Two opposing hypotheses exist: increased representational capacity might allow larger models to abstract away from superficial demographic markers, or conversely, it might enable them to more effectively encode and exploit these biases. 

Our results across Table \ref{table:bench-fairness-gender-query-results} and Figure \ref{fig:fairness-ci-short-query} provide little evidence that parameter count reliably predicts robustness to gender perturbations. While some model families exhibit modest stability improvements as scale increases, models of similar size behave very differently. For example, Qwen3 achieves some of the lowest rank shifts across gendered languages (e.g., 85--126 positions), whereas BGE-m3 (568M) exhibits stronger directional bias (e.g., $\text{RBC} = -0.52$ in German). At the opposite end of the spectrum, the highly compact 33M-parameter Arctic Embed (s) demonstrates minimal directional bias on short queries ($|\text{RBC}| \leq 0.08$ in gendered languages), though its English-focused training warrants caution against interpreting this stability as deep morphological comprehension. Ultimately, architectural design and training data distribution appear to matter more for fairness outcomes than raw parameter scale.

\textbf{Query Length Sensitivity.} The impact of query length on demographic bias is highly model-dependent. The E5 family degrades most severely when processing detailed briefs, demonstrating both reduced stability and amplified masculine-favoring bias (e.g., E5 (large) in French shifts from $\text{RBC} = -0.56$ on short queries to $-0.96$ on briefs). For most other models, expanding the context window successfully mitigates absolute rank shifts (e.g., PPLX Embed in French drops from 267 to 122 positions). However, directional bias fluctuates unpredictably: PPLX Embed and Arctic Embed (v2, l) maintain relatively consistent RBC values across lengths, while Nomic Embed (v2) exacerbates its masculine bias on longer texts. Notably, Qwen3 reverses its polarity in Romance languages, shifting from slightly favoring feminine variants on short queries to favoring masculine ones on longer briefs. Consequently, while added context generally stabilizes ranking volatility, its effect on directional fairness varies by model.

\begin{figure*}[p]
\centering\setlength{\tabcolsep}{0.1em} 
\begin{tabular}{ >{\centering\arraybackslash} m{0.02\textwidth} m{0.5\textwidth} m{0.435\textwidth}}
    \raisebox{30pt}{\rotatebox[origin=c]{90}{\textbf{French}}} &
    \includegraphics[width=\linewidth, trim={0 0 0 0}, clip]{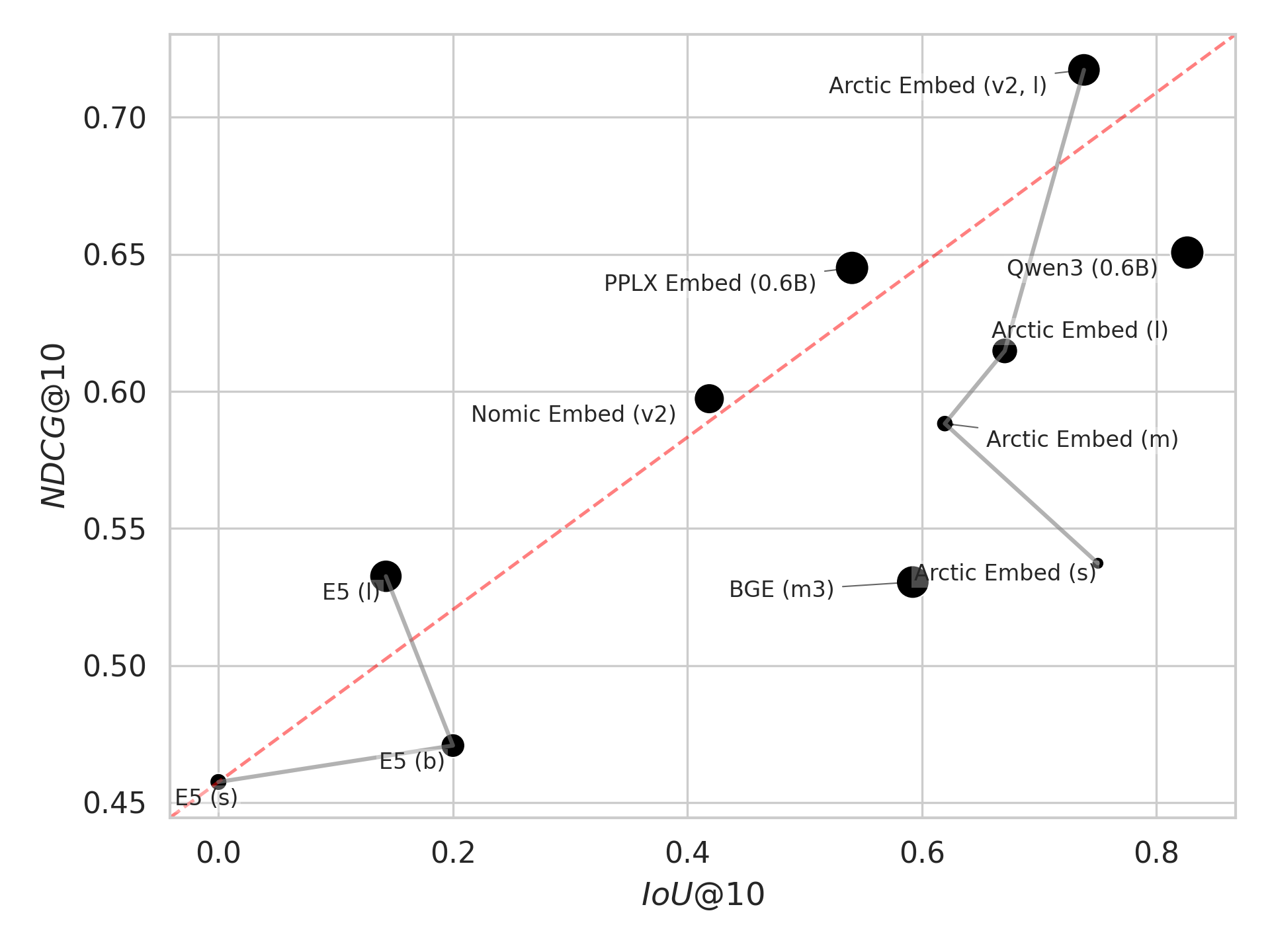} &
    \includegraphics[width=\linewidth, trim={60 0 0 0}, clip]{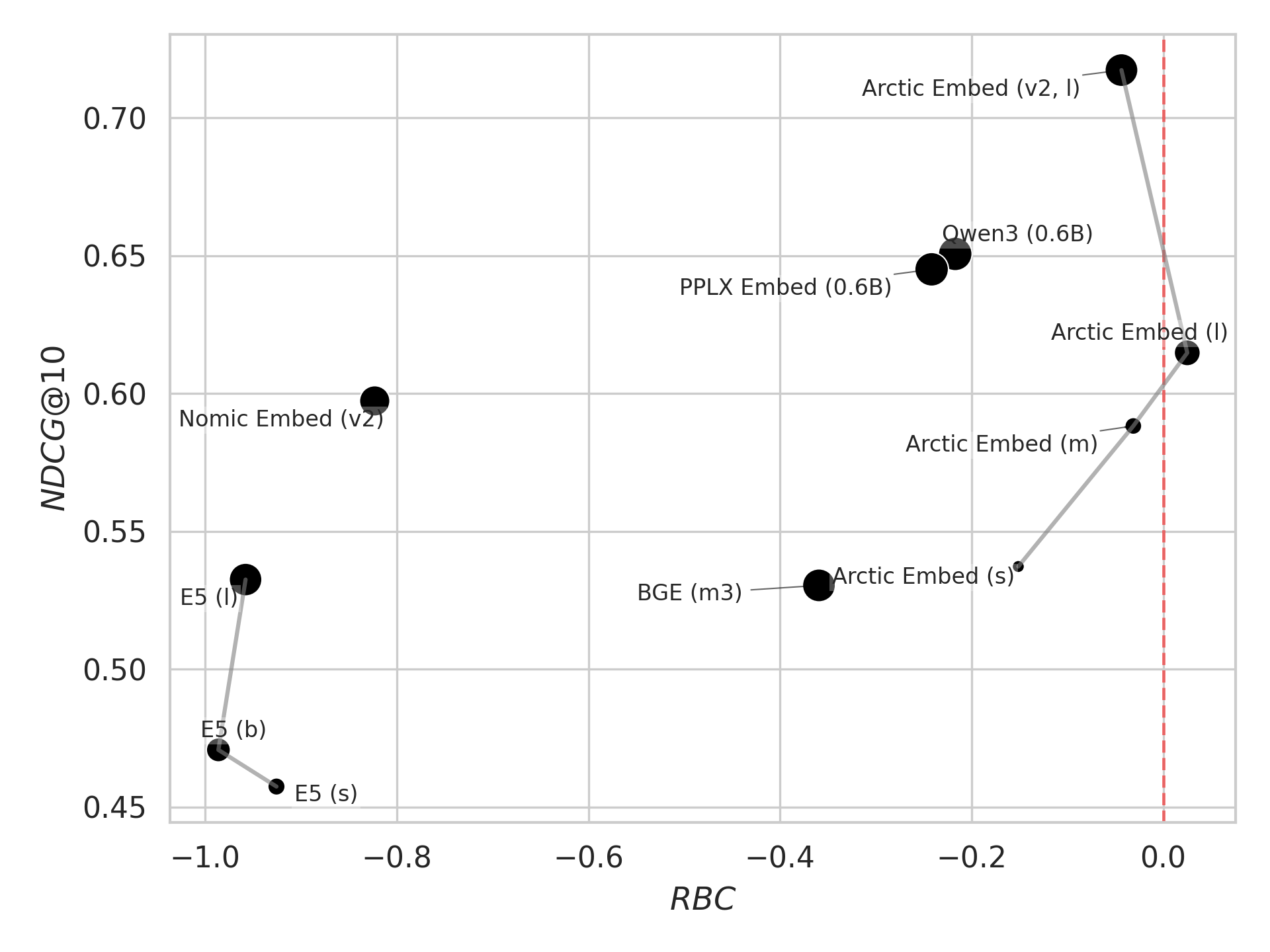} \\[-10pt]

    \raisebox{30pt}{\rotatebox[origin=c]{90}{\textbf{Spanish}}} &
    \includegraphics[width=\linewidth, trim={0 0 0 0}, clip]{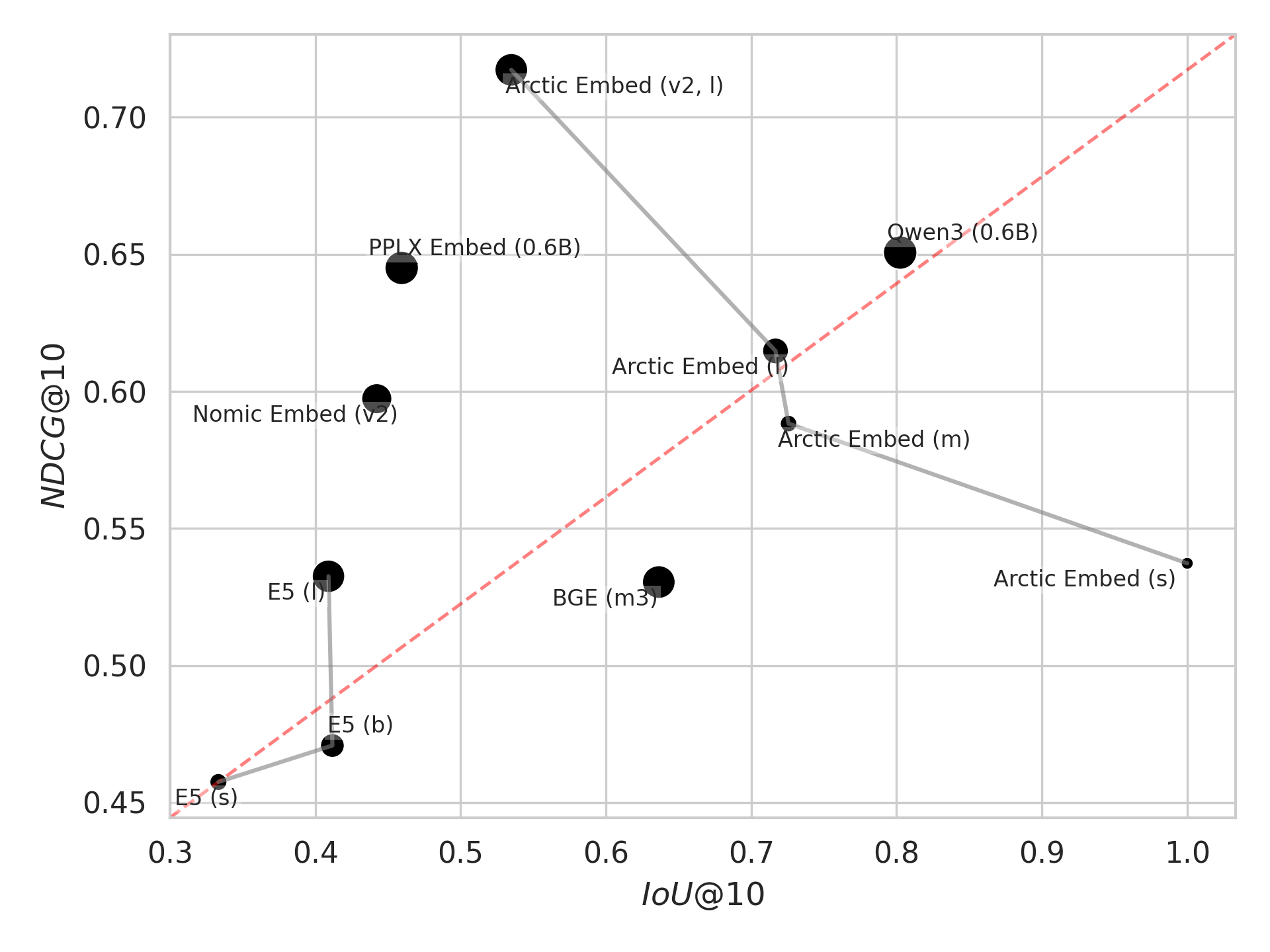} &
    \includegraphics[width=\linewidth, trim={60 0 0 0}, clip]{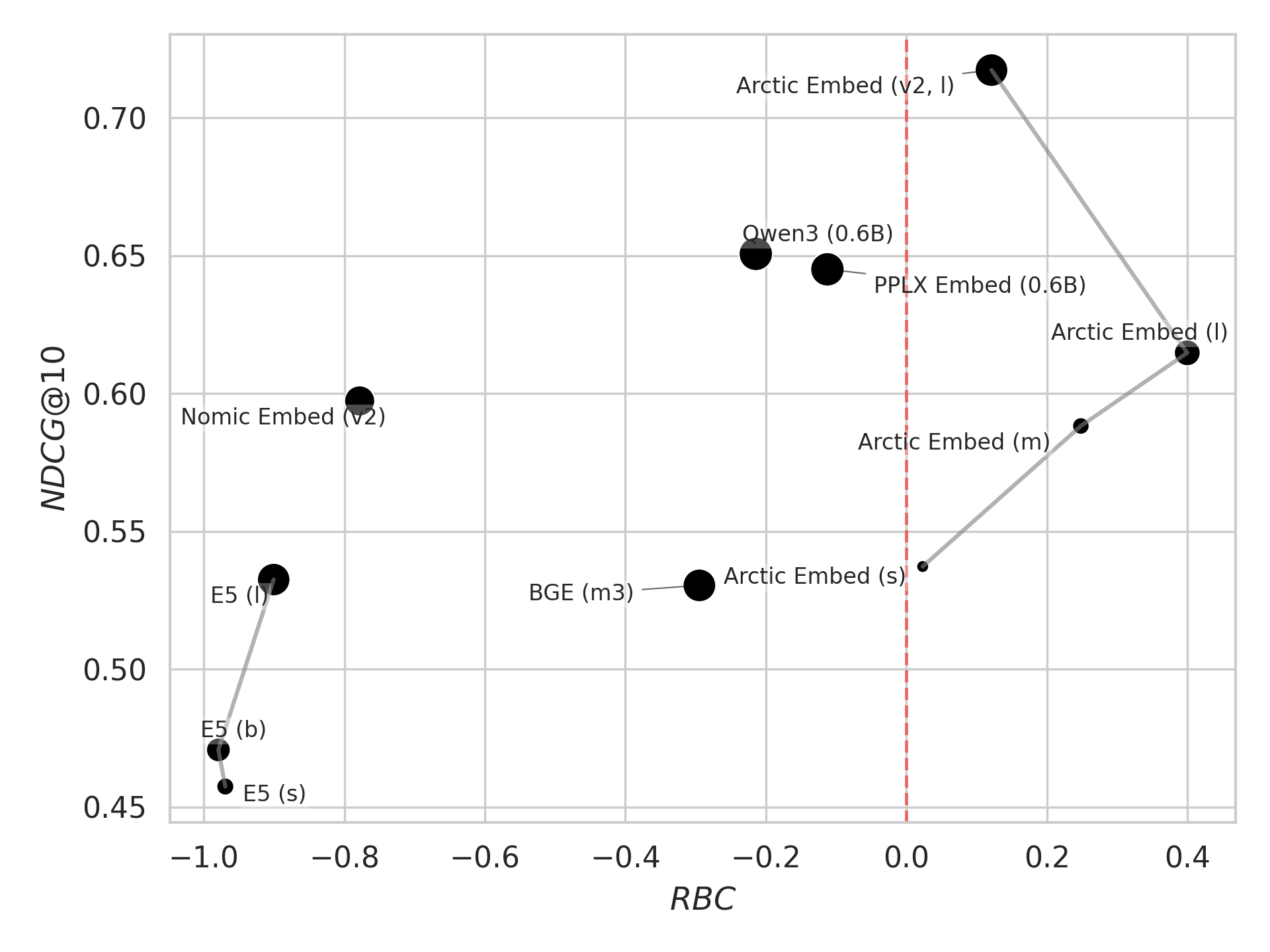} \\[-10pt]

    \raisebox{30pt}{\rotatebox[origin=c]{90}{\textbf{German}}} &
    \includegraphics[width=\linewidth, trim={0 0 0 0}, clip]{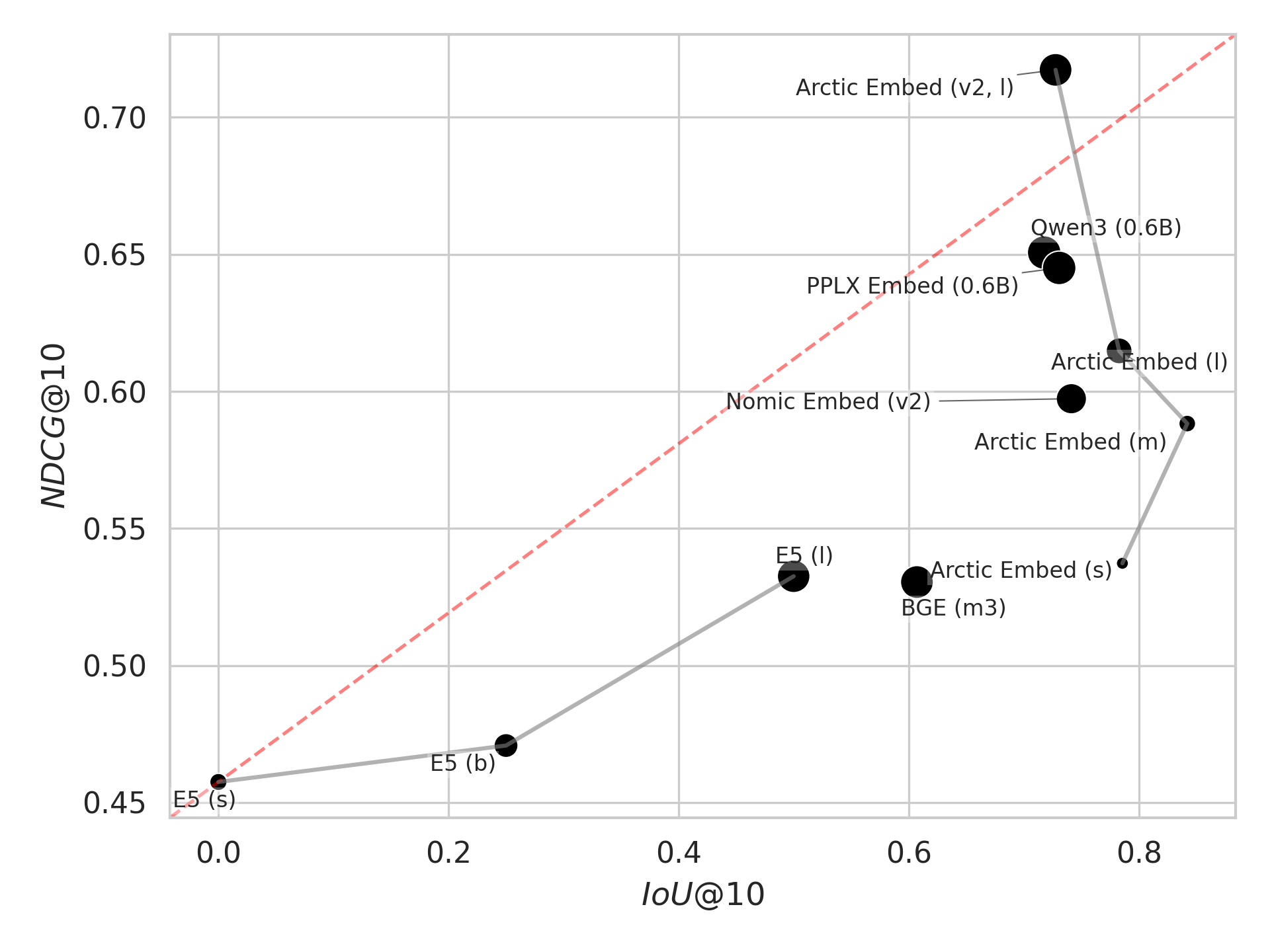} &
    \includegraphics[width=\linewidth, trim={60 0 0 0}, clip]{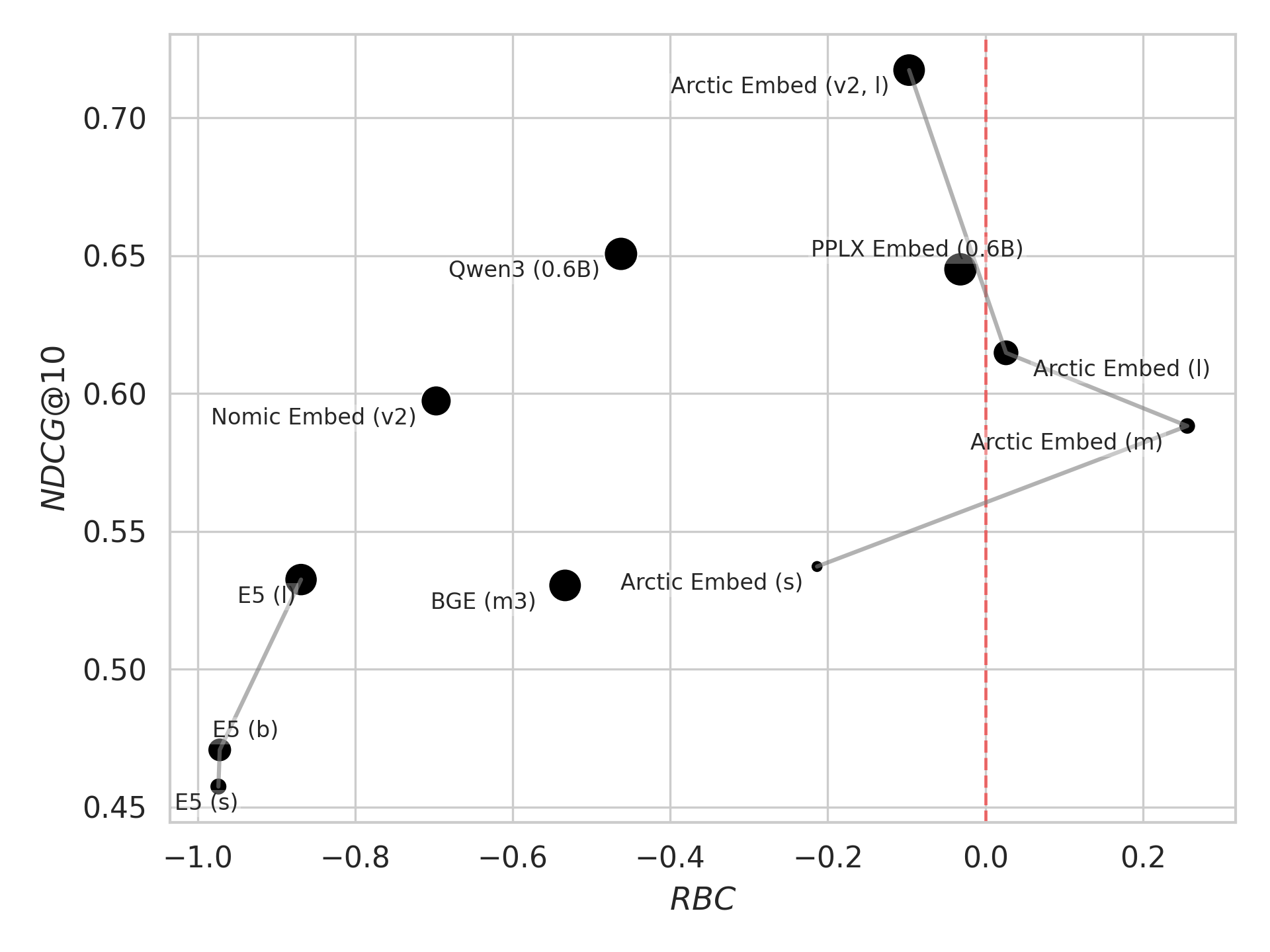} \\[-10pt]
\end{tabular}
\caption{Brief retrieval: Pareto frontiers illustrating the trade-off between multilingual retrieval effectiveness (NDCG@10) and fairness (IoU@10, RBC) per language.}
\Description{A 3 by 2 grid of scatter plots comparing retrieval effectiveness (NDCG@10) against fairness metrics across 11 embedding models on the Brief subset. The left column plots IoU@10, and the right column plots Rank-Biserial Correlation (RBC). Rows represent French, Spanish, and German. Grey lines connect the outermost models to illustrate the Pareto frontier.}
\label{fig:pareto-optimal-brief}
\end{figure*}

\begin{figure*}[p]
\centering\setlength{\tabcolsep}{0.1em} 
\begin{tabular}{ >{\centering\arraybackslash} m{0.02\textwidth} m{0.5\textwidth} m{0.435\textwidth}}
    \raisebox{30pt}{\rotatebox[origin=c]{90}{\textbf{French}}} &
    \includegraphics[width=\linewidth, trim={0 0 0 0}, clip]{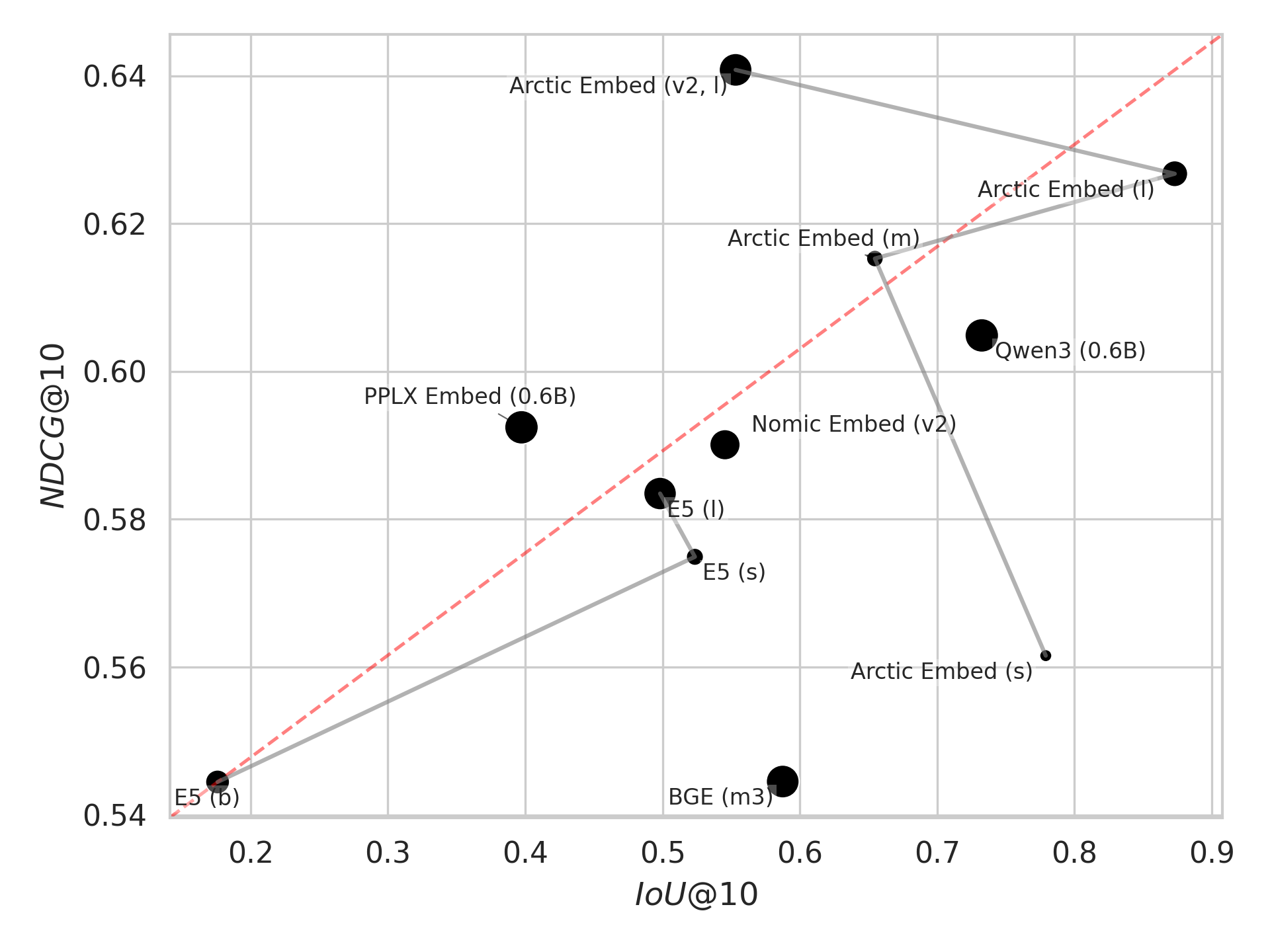} &
    \includegraphics[width=\linewidth, trim={60 0 0 0}, clip]{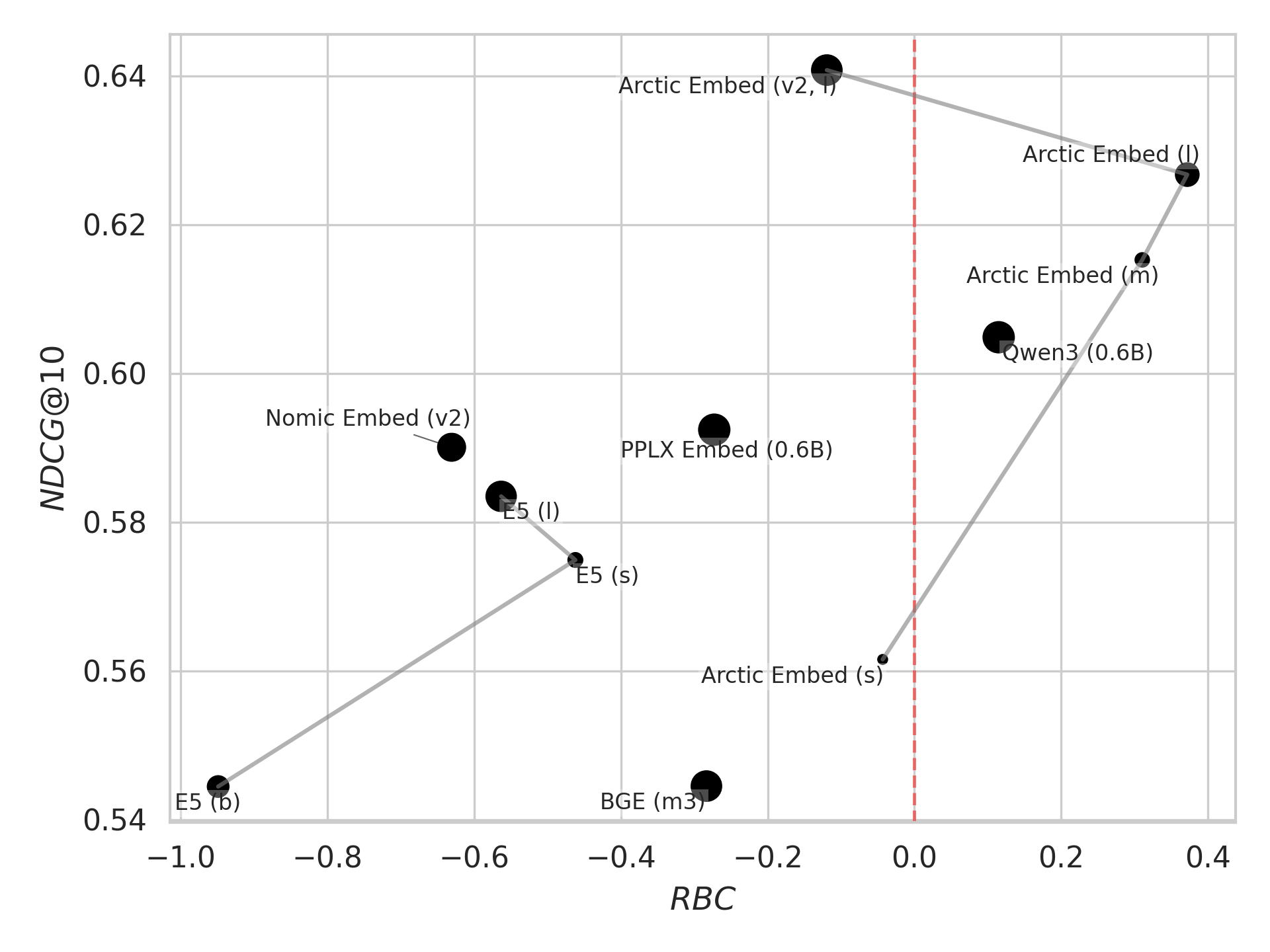} \\[-10pt]

    \raisebox{30pt}{\rotatebox[origin=c]{90}{\textbf{Spanish}}} &
    \includegraphics[width=\linewidth, trim={0 0 0 0}, clip]{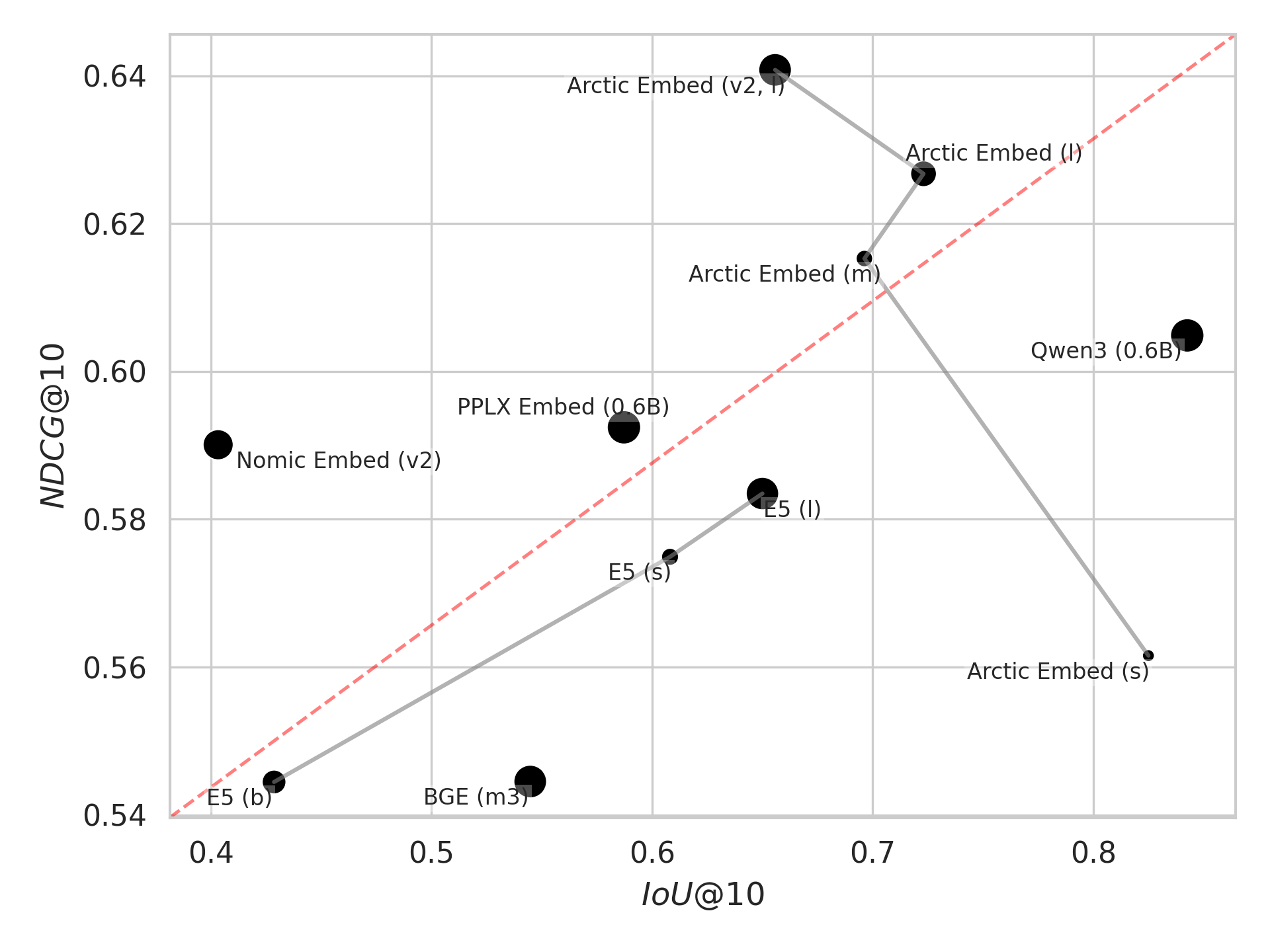} &
    \includegraphics[width=\linewidth, trim={60 0 0 0}, clip]{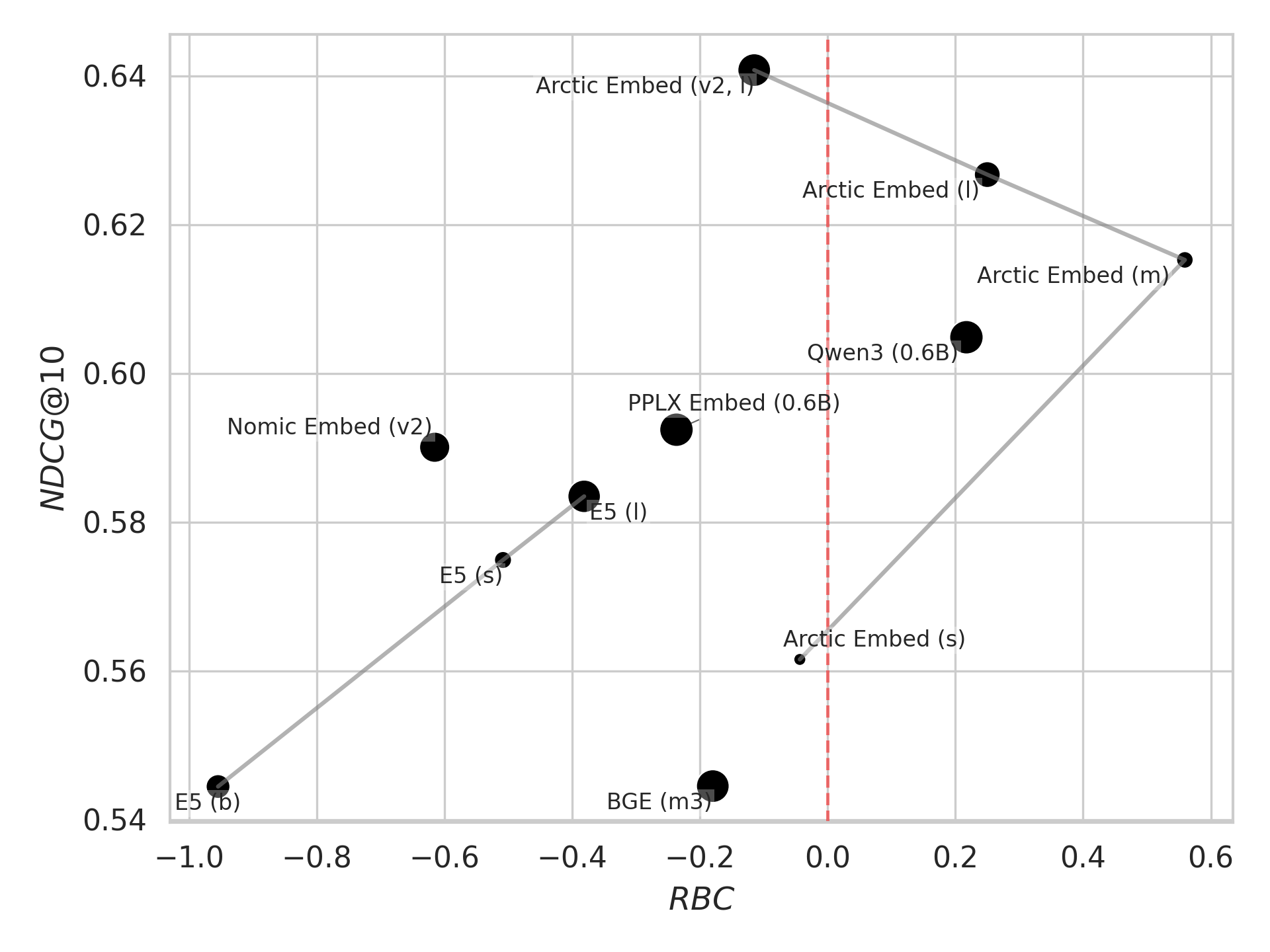} \\[-10pt]

    \raisebox{30pt}{\rotatebox[origin=c]{90}{\textbf{German}}} &
    \includegraphics[width=\linewidth, trim={0 0 0 0}, clip]{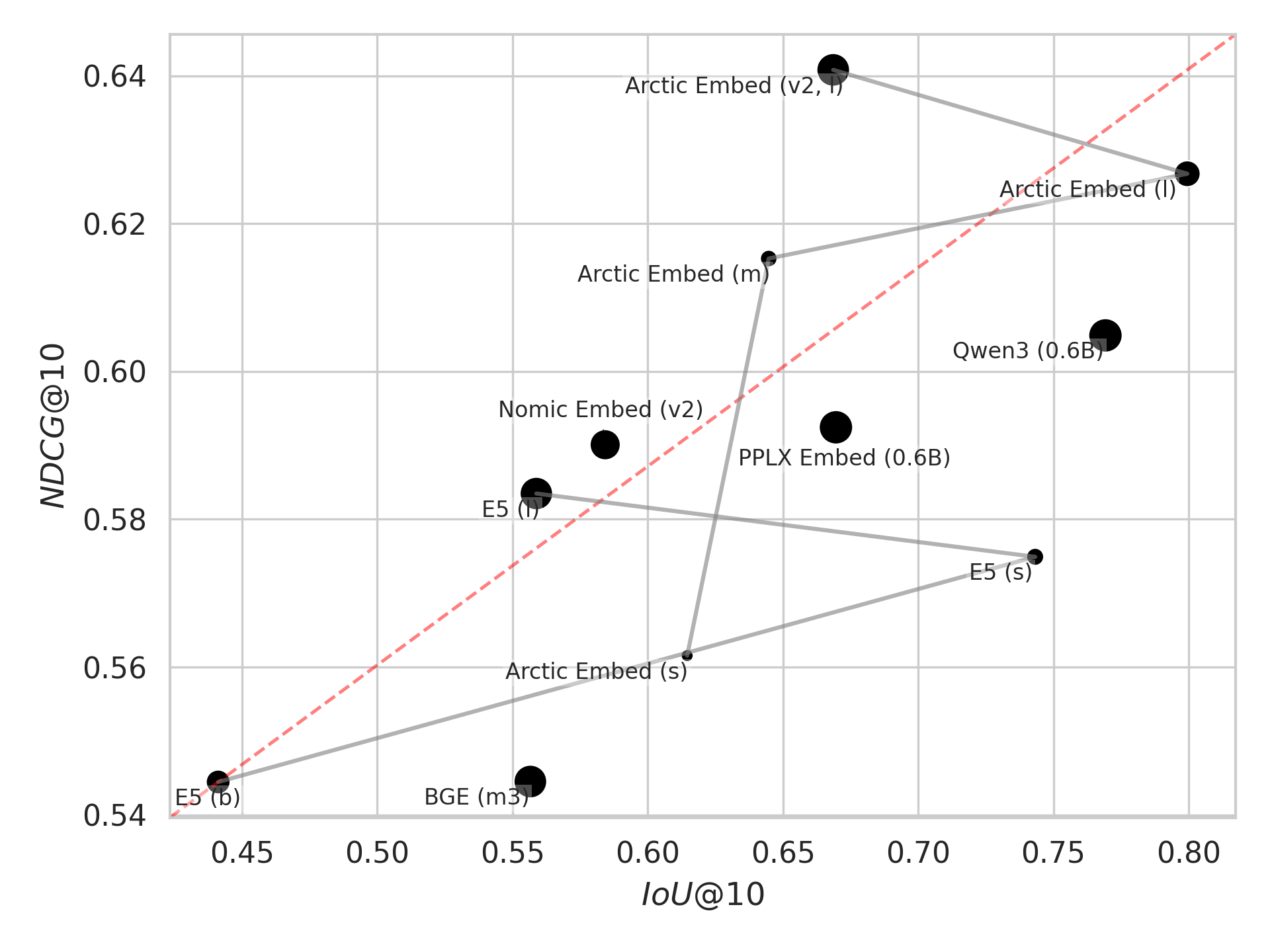} &
    \includegraphics[width=\linewidth, trim={60 0 0 0}, clip]{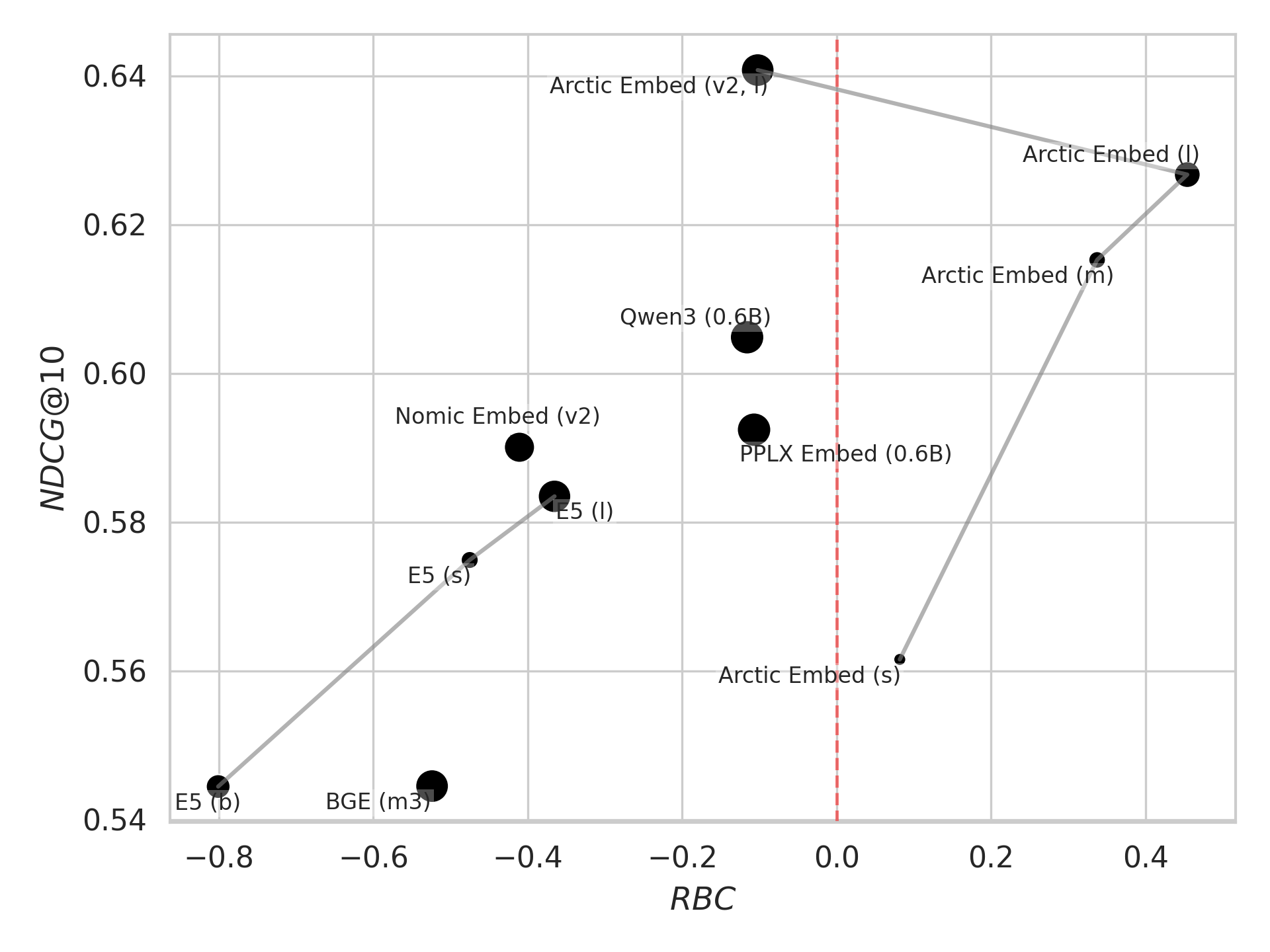} \\[-10pt]
\end{tabular}
\caption{Short Query retrieval: Pareto frontiers illustrating the trade-off between multilingual retrieval effectiveness (NDCG@10) and fairness (IoU@10, RBC) per language.}
\Description{A 3 by 2 grid of scatter plots comparing retrieval effectiveness (NDCG@10) against fairness metrics across 11 embedding models on the Short Query subset. The left column plots IoU@10, and the right column plots Rank-Biserial Correlation (RBC). Rows represent French, Spanish, and German. Grey lines connect the outermost models to illustrate the Pareto frontier.}
\label{fig:pareto-optimal-query}
\end{figure*}

\textbf{Efficiency-Fairness Trade-off.} To systematically identify models that balance retrieval accuracy with demographic robustness, Figures \ref{fig:pareto-optimal-brief} and \ref{fig:pareto-optimal-query} map retrieval effectiveness (NDCG@10) against stability (IoU@10) and directional disparity (RBC). English and Dutch are omitted from these grids, as all models achieved perfect stability ($\text{IoU}@10 = 1.00$) in those languages. 

A model is considered Pareto-optimal if no alternative achieves strictly superior performance across both axes. In the IoU--NDCG plots, the optimal frontier lies in the upper-right quadrant (maximizing both accuracy and stability). In the RBC--NDCG plots, optimal models push toward the upper center (maximizing accuracy while driving RBC toward zero). 

On briefs, Arctic Embed (v2, l) leads the effectiveness frontier across all languages due to its dominant NDCG@10. However, its fairness profile is nuanced: it maintains low directional bias ($|\text{RBC}| \leq 0.12$) but suffers from moderate instability in Spanish ($\text{IoU}@10 = 0.53$), where Qwen3 and Arctic Embed (s) offer superior stability at the cost of retrieval performance. On short queries, the original Arctic Embed (l) dominates the effectiveness frontier and leads stability in French and German, while the v2 variant remains Pareto-optimal regarding RBC. Conversely, the E5 family is dominated across nearly all configurations, yielding lower effectiveness and poorer fairness. Overall, Arctic Embed (v2, l) represents the most compelling production trade-off currently available. Beyond these specific models, this dual-axis evaluation framework lets practitioners to map their own proprietary systems onto the effectiveness-fairness frontier to navigate application-specific trade-offs.

%% file: tables/gender/query/retrieval-fairness.tex
\begin{table*}[ht]
\caption{{Gendered form retrieval fairness results on Short Query.\\\footnotesize\textit{ \colorbox[HTML]{DBDBDB}{\textbf{Best}} and \colorbox[HTML]{F0F0F0}{\textbf{Second Best}} values are highlighted per column.}}}
\centering\footnotesize\setlength{\tabcolsep}{1pt}
\begin{tabularx}{\textwidth}{*{1}{l}*{3}{>{\centering\arraybackslash}X}*{3}{>{\centering\arraybackslash}X}*{3}{>{\centering\arraybackslash}X}*{3}{>{\centering\arraybackslash}X}*{3}{>{\centering\arraybackslash}X}}
\toprule
\bfseries Model & \multicolumn{3}{c}{\bfseries English} & \multicolumn{3}{c}{\bfseries French} & \multicolumn{3}{c}{\bfseries Spanish} & \multicolumn{3}{c}{\bfseries German} & \multicolumn{3}{c}{\bfseries Dutch} \\
 \cmidrule(lr){1-1} \cmidrule(lr){2-4} \cmidrule(lr){5-7} \cmidrule(lr){8-10} \cmidrule(lr){11-13} \cmidrule(lr){14-16}
Name (version) & IoU @10 & RBC & Rank Shift & IoU @10 & RBC & Rank Shift & IoU @10 & RBC & Rank Shift & IoU @10 & RBC & Rank Shift & IoU @10 & RBC & Rank Shift \\
\midrule
Arctic Embed (s) & \bfseries {\cellcolor[HTML]{DBDBDB}} 1.00 & \bfseries {\cellcolor[HTML]{F0F0F0}} -0.01 & \bfseries {\cellcolor[HTML]{F0F0F0}} 0.00 & \bfseries {\cellcolor[HTML]{F0F0F0}} 0.78 & \bfseries {\cellcolor[HTML]{DBDBDB}} -0.04 & 147 & \bfseries {\cellcolor[HTML]{F0F0F0}} 0.82 & \bfseries {\cellcolor[HTML]{DBDBDB}} -0.04 & 130 & 0.61 & \bfseries {\cellcolor[HTML]{DBDBDB}} 0.08 & 176 & \bfseries {\cellcolor[HTML]{DBDBDB}} 1.00 & 0.01 & \bfseries {\cellcolor[HTML]{DBDBDB}} 0.00 \\
E5 (s) & \bfseries {\cellcolor[HTML]{DBDBDB}} 1.00 & -0.88 & 0.23 & 0.52 & -0.46 & 216 & 0.61 & -0.51 & 229 & 0.74 & -0.48 & 227 & \bfseries {\cellcolor[HTML]{DBDBDB}} 1.00 & 0.01 & \bfseries {\cellcolor[HTML]{DBDBDB}} 0.00 \\
Arctic Embed (m) & \bfseries {\cellcolor[HTML]{DBDBDB}} 1.00 & \bfseries {\cellcolor[HTML]{DBDBDB}} 0.00 & \bfseries {\cellcolor[HTML]{F0F0F0}} 0.00 & 0.65 & 0.31 & 172 & 0.70 & 0.56 & 221 & 0.64 & 0.34 & 165 & \bfseries {\cellcolor[HTML]{DBDBDB}} 1.00 & \bfseries {\cellcolor[HTML]{DBDBDB}} 0.00 & \bfseries {\cellcolor[HTML]{DBDBDB}} 0.00 \\
E5 (b) & \bfseries {\cellcolor[HTML]{DBDBDB}} 1.00 & -0.15 & 0.14 & 0.18 & -0.95 & 430 & 0.43 & -0.95 & 390 & 0.44 & -0.80 & 293 & \bfseries {\cellcolor[HTML]{DBDBDB}} 1.00 & 0.01 & \bfseries {\cellcolor[HTML]{DBDBDB}} 0.00 \\
Nomic Embed (v2) & \bfseries {\cellcolor[HTML]{DBDBDB}} 1.00 & -0.04 & 0.02 & 0.55 & -0.63 & 277 & 0.40 & -0.62 & 330 & 0.58 & -0.41 & 272 & \bfseries {\cellcolor[HTML]{DBDBDB}} 1.00 & \bfseries {\cellcolor[HTML]{DBDBDB}} 0.00 & \bfseries {\cellcolor[HTML]{DBDBDB}} 0.00 \\
E5 (l) & \bfseries {\cellcolor[HTML]{DBDBDB}} 1.00 & -0.32 & 0.08 & 0.50 & -0.56 & 252 & 0.65 & -0.38 & 216 & 0.56 & -0.37 & 210 & \bfseries {\cellcolor[HTML]{DBDBDB}} 1.00 & -0.04 & \bfseries {\cellcolor[HTML]{DBDBDB}} 0.00 \\
Arctic Embed (l) & \bfseries {\cellcolor[HTML]{DBDBDB}} 1.00 & \bfseries {\cellcolor[HTML]{DBDBDB}} 0.00 & \bfseries {\cellcolor[HTML]{DBDBDB}} 0.00 & \bfseries {\cellcolor[HTML]{DBDBDB}} 0.87 & 0.37 & \bfseries {\cellcolor[HTML]{F0F0F0}} 137 & 0.72 & 0.25 & \bfseries {\cellcolor[HTML]{F0F0F0}} 124 & \bfseries {\cellcolor[HTML]{DBDBDB}} 0.80 & 0.45 & \bfseries {\cellcolor[HTML]{F0F0F0}} 161 & \bfseries {\cellcolor[HTML]{DBDBDB}} 1.00 & \bfseries {\cellcolor[HTML]{F0F0F0}} -0.01 & \bfseries {\cellcolor[HTML]{DBDBDB}} 0.00 \\
Arctic Embed (v2, l) & \bfseries {\cellcolor[HTML]{DBDBDB}} 1.00 & 0.06 & 0.05 & 0.55 & -0.12 & 179 & 0.66 & \bfseries {\cellcolor[HTML]{F0F0F0}} -0.12 & 211 & 0.67 & \bfseries {\cellcolor[HTML]{F0F0F0}} -0.10 & 190 & \bfseries {\cellcolor[HTML]{DBDBDB}} 1.00 & \bfseries {\cellcolor[HTML]{DBDBDB}} 0.00 & \bfseries {\cellcolor[HTML]{DBDBDB}} 0.00 \\
BGE (m3) & \bfseries {\cellcolor[HTML]{DBDBDB}} 1.00 & 0.04 & 0.07 & 0.59 & -0.28 & 189 & 0.54 & -0.18 & 220 & 0.56 & -0.52 & 252 & \bfseries {\cellcolor[HTML]{DBDBDB}} 1.00 & \bfseries {\cellcolor[HTML]{F0F0F0}} -0.01 & \bfseries {\cellcolor[HTML]{DBDBDB}} 0.00 \\
PPLX Embed (0.6B) & \bfseries {\cellcolor[HTML]{DBDBDB}} 1.00 & 0.27 & 0.25 & 0.40 & -0.27 & 267 & 0.59 & -0.24 & 232 & 0.67 & -0.11 & 190 & \bfseries {\cellcolor[HTML]{DBDBDB}} 1.00 & \bfseries {\cellcolor[HTML]{F0F0F0}} -0.01 & \bfseries {\cellcolor[HTML]{DBDBDB}} 0.00 \\
Qwen3 (0.6B) & \bfseries {\cellcolor[HTML]{DBDBDB}} 1.00 & 0.99 & 0.63 & 0.73 & \bfseries {\cellcolor[HTML]{F0F0F0}} 0.11 & \bfseries {\cellcolor[HTML]{DBDBDB}} 116 & \bfseries {\cellcolor[HTML]{DBDBDB}} 0.84 & 0.22 & \bfseries {\cellcolor[HTML]{DBDBDB}} 85 & \bfseries {\cellcolor[HTML]{F0F0F0}} 0.77 & -0.12 & \bfseries {\cellcolor[HTML]{DBDBDB}} 126 & \bfseries {\cellcolor[HTML]{DBDBDB}} 1.00 & \bfseries {\cellcolor[HTML]{DBDBDB}} 0.00 & \bfseries {\cellcolor[HTML]{DBDBDB}} 0.00 \\
\bottomrule
\end{tabularx}

\label{table:bench-fairness-gender-query-results}
\end{table*}